\documentclass[runningheads]{llncs}

\usepackage{eccv}

\usepackage{eccvabbrv}

\usepackage{graphicx}
\usepackage{booktabs}

\usepackage[accsupp]{axessibility}  
\usepackage{pifont}
\usepackage{multirow}
\usepackage{makecell}
\usepackage{wrapfig}

\newcommand{\dataset}{M-BAPPS\xspace}
\newcommand{\model}{DepictQA\xspace}
\newcommand{\supp}{\textit{Appendix}\xspace}
\definecolor{darkred}{rgb}{0.7, 0.0, 0.0}
\definecolor{darkgreen}{rgb}{0.0, 0.37, 0.14}
\definecolor{darkblue}{rgb}{0.10, 0.17, 0.8}
\definecolor{lightblue}{RGB}{37, 150, 190}

\usepackage[breaklinks,colorlinks,citecolor=eccvblue]{hyperref}
\usepackage{orcidlink}

\begin{document}

\title{
\mbox{Depicting Beyond Scores:}
\mbox{Advancing Image Quality Assessment through}
\mbox{Multi-modal Language Models}
}

\titlerunning{DepictQA: Depicted Image Quality Assessment}

\author{
\mbox{
Zhiyuan You\inst{12}$^*$\orcidlink{0009-0006-8546-3478} \and
Zheyuan Li\inst{26}$^*$\orcidlink{0009-0005-9078-843X} \and
Jinjin Gu\inst{34}$^*$\orcidlink{0000-0002-4389-6236} \and
Zhenfei Yin\inst{34}\orcidlink{0000-0002-8666-1103} \and}
\mbox{Tianfan Xue\inst{1}$^\dag$\orcidlink{0000-0001-5031-6618} \and Chao Dong\inst{235}$^\dag$\orcidlink{0000-0003-2260-8079}}
}

\authorrunning{Z. You et al.}

\institute{The Chinese University of Hong Kong \and 
Shenzhen Institute of Advanced Technology, Chinese Academy of Sciences \and 
\mbox{Shanghai AI Laboratory \quad \and University of Sydney} \and
\mbox{Shenzhen University of Advanced Technology \quad \and University of Macau}
\email{zhiyuanyou@foxmail.com  zheyuanli884886@gmail.com  jinjin.gu@sydney.edu.au}
\email{zyin7056@uni.sydney.edu.au  tfxue@ie.cuhk.edu.hk  chao.dong@siat.ac.cn}
\mbox{$^*$ Contribute Equally \quad $^\dag$ Corresponding Author}
\mbox{Project Page: \url{https://depictqa.github.io}}
}

\maketitle

\begin{abstract}

We introduce a \textbf{Depict}ed image \textbf{Q}uality \textbf{A}ssessment method (\model), overcoming the constraints of traditional score-based methods. 
\model allows for detailed, language-based, human-like evaluation of image quality by leveraging Multi-modal Large Language Models (MLLMs). 
Unlike conventional Image Quality Assessment (IQA) methods relying on scores, \model interprets image content and distortions descriptively and comparatively, aligning closely with humans' reasoning process. 
To build the \model model, we establish a hierarchical task framework, and collect a multi-modal IQA training dataset. 
To tackle the challenges of limited training data and multi-image processing, we propose to use multi-source training data and specialized image tags. 
These designs result in a better performance of \model than score-based approaches on multiple benchmarks. 
Moreover, compared with general MLLMs, \model can generate more accurate reasoning descriptive languages. 
We also demonstrate that our full-reference dataset can be extended to non-reference applications. 
These results showcase the research potential of multi-modal IQA methods.


\end{abstract}

\section{Introduction}\label{sec:introduction}

\definecolor{color_distortion}{rgb}{0.02, 0.20, 1.00}
\definecolor{color_textue}{rgb}{0.58, 0.13, 0.57}
\definecolor{color_conclusion}{rgb}{0.0, 0.56, 0.32}
\begin{figure}[t]
    \centering
    \includegraphics[width=0.95\linewidth]{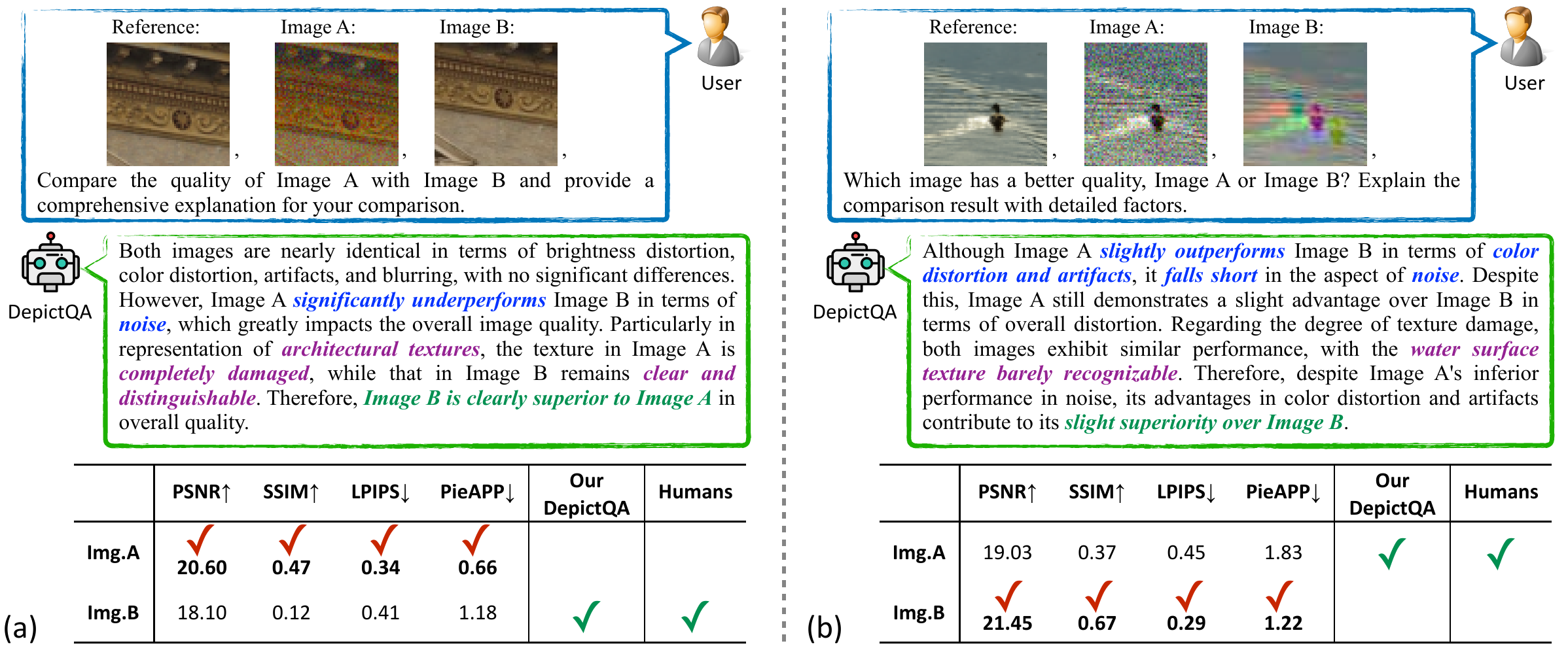}
    \caption{
        \textbf{Comparison} between our \model and score-based IQA methods, including PSNR, SSIM~\cite{ssim}, LPIPS~\cite{bapps}, and PieAPP~\cite{pieapp}.
        Score-based IQA methods only provide numerical scores devoid of reasoning and justification.
        Thus they disagree with human judgments in complex scenarios when \textbf{(a)} images are misaligned and \textbf{(b)} both images suffer from severe distortions. 
        In contrast, \model first identifies the \textcolor{color_distortion}{distortions} of images, then weighs the influences of different distortions to the \textcolor{color_textue}{texture damages}, and finally obtains the \textcolor{color_conclusion}{comparison results} that are better aligned with human judgments. 
    }
    \label{fig:teaser}
\end{figure}

Image Quality Assessment (IQA) is an important topic in low-level vision research~\cite{MUSIQ,NIQE,bapps,gu2020image}, and it is widely applied in image generation and processing~\cite{ldm,chen2024deep,chen2023prompt,chen2023comparative}. 
IQA aims to measure and compare the quality of images, expecting the final results to be aligned with human judgments. 
Existing IQA methods~\cite{ssim,bapps,pieapp,gu2020image} mainly output the quality or similarity scores, which have apparent shortcomings. 
First, image quality is affected by different factors that cannot be effectively expressed by a simple score, \eg, noise, color distortion, and artifacts in \cref{fig:teaser}. 
Second, the reasoning process by humans cannot be well modeled by current IQA methods. 
For example, in \cref{fig:teaser}b, humans may first identify the distortions (\textit{i.e.}, noise in Image A, color distortion and artifacts in Image B), then weigh the impacts of these distortions on overall visual quality (color distortion and artifacts in Image B are worse than noise in Image A), and finally conclude that Image A is better than Image B. 
On the contrary, existing IQA methods simply compare the quality scores of these two images.

To better align with humans, we explore a new paradigm for IQA, named \textbf{Depict}ed image \textbf{Q}uality \textbf{A}ssessment (\model). 
Inspired by recent Large Language Models (LLMs)~\cite{llama, gpt4} and multi-modal technologies~\cite{llava, minigpt4}, we believe that language is the key to solving the above problems. 
As shown in \cref{fig:teaser}, \model takes both images and a question as inputs, then outputs a paragraph that describes the quality of images from multiple aspects.
Furthermore, empowered by the reasoning capability of LLMs, \model can weigh the importance of each distortion and make the final judgment. 
For instance, in \cref{fig:teaser}a, \model finds that ``the texture in Image A is completely damaged'' while ``Image B remains clear and distinguishable'', thus concludes ``Image B is superior to Image A''. 
Learning this kind of reasoning makes \model better at aligning human judgments than existing methods in complex scenarios like misalignment (\cref{fig:teaser}a) and multiple distortions (\cref{fig:teaser}b). 
Meanwhile, these descriptive outputs can be naturally understood by humans, greatly improving the interpretability.

To integrate language into IQA, we establish a hierarchy of tasks, inspired by human evaluation. 
Humans first perceive the distortions of the image, then use this information to determine the image quality. 
Also, it is easier for humans to compare the difference between two images in a single dimension (\eg, color distortion) than quantitatively evaluate the overall quality of an image or the similarity between two images, as verified by~\cite{ponomarenko2009tid2008, tid2013, pieapp}. 
Based on this intuition, \model does not produce scores, but describes the image quality and compares two images.
Specifically, we break \model task into a hierarchy of 3 tasks (detailed in \cref{fig:dataset}): (1) Quality Description, (2) Quality Comparison, and (3) Comparison Reasoning. 
These designs follow the process of human evaluation.

To train the proposed \model, we further construct a multi-modal IQA dataset, named \dataset, by collecting text descriptions based on the existing BAPPS IQA dataset~\cite{bapps}. 
Our \dataset dataset contains 5,104 detailed high-quality text descriptions and 115,646 brief descriptions. 
For high-quality texts, we first collect the quality-related information through a carefully designed questionnaire (details shown in \cref{fig:dataset}a and \ref{fig:dataset}c), the results of which are further converted into a descriptive paragraph using GPT-4~\cite{gpt4}. 
To further increase the size of the training set, we also augment the dataset with brief descriptions. 
Specifically, we convert the existing quality comparison label in BAPPS into a brief description using pre-generated templated texts, such as ``Image A maintains a distinct advantage in terms of image quality over Image B''.

With the dataset mentioned above, we resort to Multi-modal Large Language Models (MLLMs)~\cite{llava, minigpt4, kosmos} to bridge the gap between images and descriptive texts. 
However, directly applying existing MLLMs to our \model faces two challenges. 
First, there are only limited images with high-quality descriptions, preventing the model from robustly correlating images and text descriptions. 
In this aspect, we present a multi-source training approach to increase the size of training data. 
Specifically, two additional sources are used. 
One is images with only brief templated texts, as mentioned above. 
The other one is external quality-unrelated content description data, the Detailed Description dataset in~\cite{lamm}, which contains 48,734 image-text pairs. 
Although these two datasets are not directly designed for the descriptive reasoning ability, we find that the former one can help bridge images and texts in quality-related tasks, while the latter one can serve as a regularization. 
Second, many MLLMs have difficulty in distinguishing multiple images, but our setup requires two or more images. 
We solve this problem by employing specialized tags for different images, instead of a unified tag for all images. 
Empirical results demonstrate that these approaches effectively mitigate the two challenges and bring a better \model model.

Finally, we conduct extensive experiments to prove the effectiveness of our \model.
First, \model achieves state-of-the-art performance on multiple existing IQA benchmark, well aligned with human judgments.
Also, \model can describe the distortions and texture damages in images and explain the reasoning process when comparing two images, thus generating more accurate descriptions compared with general-purpose MLLMs. 
Even compared with notably GPT-4V~\cite{gpt4v}, \model has significantly better comparison ability and comparable reasoning ability. 
Moreover, we demonstrate the utility of our full-reference dataset in non-reference applications. 
These results attest to the superiority of our \model and the research potential of multi-modal IQA tasks.

\section{Related Works}\label{sec:related_works}

\textbf{Score-based IQA methods}. 
Most existing IQA methods rely on scores to assess image quality. 
They can be categorized into \textit{full-reference} and \textit{non-reference} methods. 
(1) Full-reference methods assess image quality by computing the similarity score between a distorted image and a high-quality reference image. 
Traditional methods rely on human-designed metrics like structural similarity~\cite{ssim}, image information~\cite{vif}, phase congruency with gradient magnitude~\cite{fsim}, \textit{etc}. 
Learning-based methods aim to align with human assessment through data-driven training.
LPIPS~\cite{bapps} shows that the learned features can effectively function as a perceptual metric, exhibiting high consistency with human judgments.
In alignment with advancements in the deep-learning community, data-driven approaches~\cite{pieapp, WaDIQaM, dists, A-DISTS, JSPL, CVRKD, SRIF, ghildyal2022stlpips} have similarly spurred innovations in IQA. 
(2) Non-reference methods evaluate the quality of a distorted image without a reference image. 
Traditional methods~\cite{moorthy2010two, DIIVINE, NIQE, BRISQUE, MA, saad2012blind, tang2011learning} primarily calculate quality scores based on human-designed natural image statistics.
Deep-learning-based methods~\cite{CNNIQA, RankIQA, BPSQM, MetaIQA, HyperIQA, MUSIQ, CKDN} replace hand-crafted statistics by learning quality priors from extensive data. 
Recent works further enhance the performance by introducing graph representation~\cite{graphiqa}, CLIP pre-training~\cite{CLIP-IQA}, continual learning~\cite{zhang2022continual}, multitask learning~\cite{LIQE}, and so on. 
However, score-based IQA methods exhibit inherent limitations, particularly the inability to reflect the intricate analyses and weights of multiple aspects, as discussed in \cref{sec:introduction}.

\textbf{MLLMs} incorporate the vision modality into large language models~\cite{vicuna, gpt4, llama}, aiming to leverage their emergent ability to achieve general vision ability. 
These MLLMs~\cite{flamingo, instructblip, llava, gpt4v, mplug-owl, lamm, zhang2023internlm, li2023blip, zhang2023llama, minigpt4} have demonstrated a general visual ability and can tackle various multi-modality tasks, including image captioning~\cite{nocaps, cococap, flickr}, visual question answering~\cite{vqav2, mmbench, scienceqa}, document understanding~\cite{chartqa, docvqa, textvqa}, \etc
Although proficient in these high-level perception tasks, we demonstrate in \cref{subsec:compare_mllm} that general MLLMs are still not good at IQA tasks.

\textbf{MLLM-based IQA methods} aim to achieve better alignment with human perception leveraging languages~\cite{iqasurvey}. 
Q-Bench~\cite{qbench, qbench_plus} constructs a benchmark to assess existing MLLMs in low-level perception tasks. 
Q-Instruct~\cite{qinstruct} and Co-Instruct~\cite{co_instruct} further promote the low-level perception ability of MLLMs by introducing large-scale datasets. 
Q-Align~\cite{qalign} utilizes the text-guided instruction tuning for more accurate quality score regression. 
Our work distinguishes itself from existing works. 
Our focus lies on quality comparison regarding distortions and texture damages across multiple images, whereas existing works primarily center on low-level perception and score regression within individual images.

\section{\model Task and Dataset}\label{sec:dataset}

\subsection{Task Description}

Before introducing our method, we need to rethink the paradigm of IQA. 
To reflect the human process of assessing image quality, we intend to apply language as a powerful interactive tool. 
Intuitively, \model needs the following abilities.

\begin{figure*}[t]
    \centering
    \includegraphics[width=\linewidth]{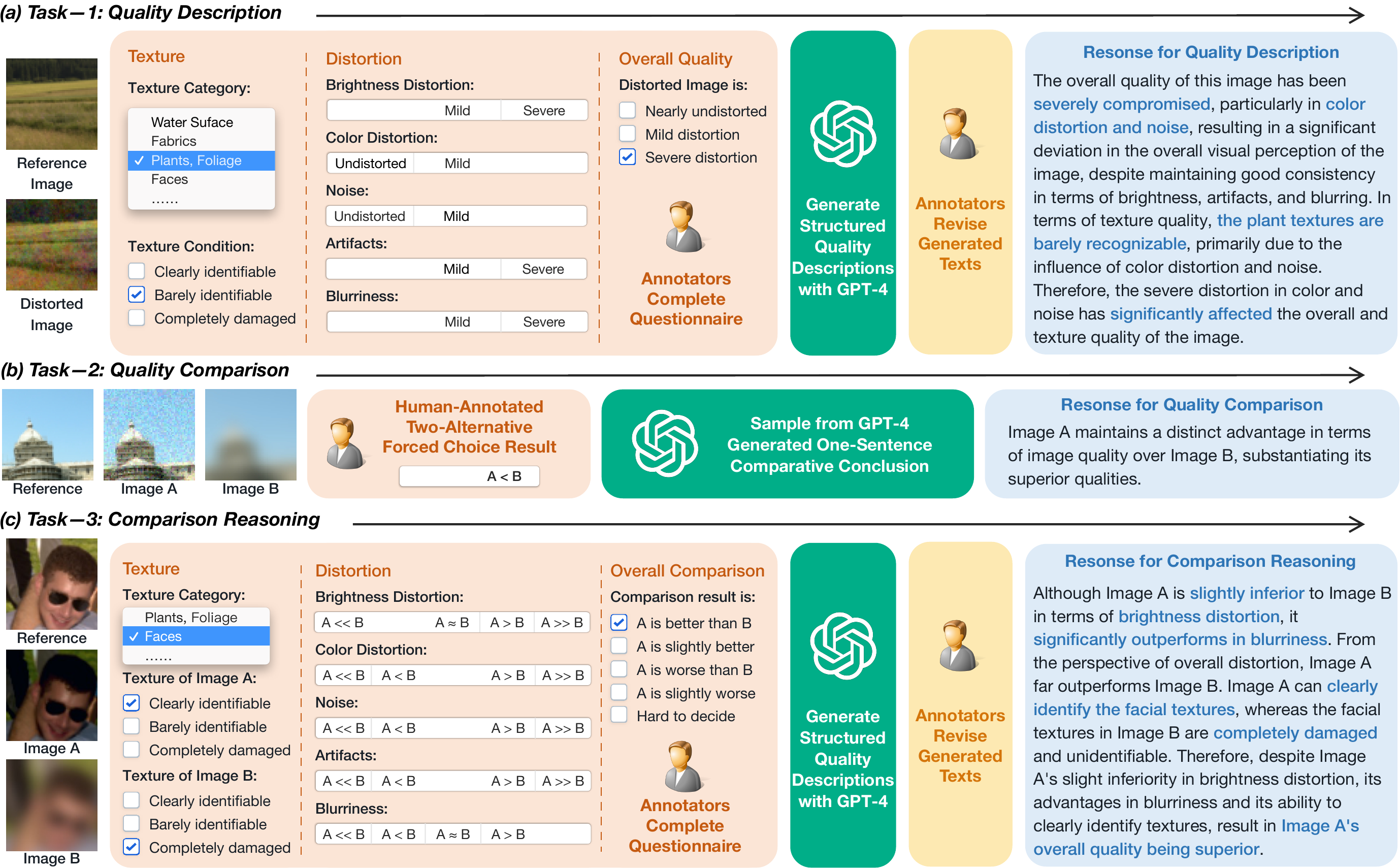}
    \caption{
        \textbf{Collection of the responses in our \dataset dataset}.
        We first carefully design a questionnaire to collect quality-related information. 
        We then employ the GPT-4~\cite{gpt4} to convert our annotated questionnaire results into natural language.
        Finally, the outputs of GPT-4 are modified and improved by the annotators to correct errors, eliminate ambiguities, and supplement important information. 
    }
    \label{fig:dataset}
\end{figure*}

First, \model needs to \textbf{identify the distortions and texture damages} (\cref{fig:dataset}a). 
Humans begin assessing image quality by identifying distortions and texture damages, as this is the basis for any subsequent assessment.

Second, \model is required to \textbf{compare distorted images} like \cref{fig:dataset}b rather than just calculate scores for individual images. 
Quantifying image quality has drawbacks, as the information from a single score is quite limited. 
It has also been verified that humans tend to make a biased quality assessment on a single image, but are more consistent and reliable in comparing two images~\cite{gu2020image,gu2020pipal,tid2013}.

Third, \model should \textbf{weigh and judge multiple aspects} that affect image quality. 
Humans consider many factors when comparing image quality. 
For example, when comparing an underexposed image and a blurry image in \cref{fig:dataset}c, one may need to consider the real impacts of these two distortions on the texture representation, and weigh among these considerations. 
\model should mimic this weighing ability, which distinguishes it from previous IQA methods.

Based on the above discussion, as described in \cref{fig:dataset}, we design a hierarchical task paradigm, progressively guiding \model to obtain the above abilities: 
\begin{itemize}
    \item \textit{Task--1: Quality Description}. 
    Based on the provided reference image and a distorted image, describe the distortions and texture damages in the distorted image, and comment on the overall quality of the distorted image. 
    \item \textit{Task--2: Quality Comparison}. 
    Based on the provided reference image and its two distorted versions, determine which distorted image has a better quality.
    \item \textit{Task--3: Comparison Reasoning}. 
    Based on the comparison result, describe the distortions and texture damages of the two distorted images, make inferences, and weigh the pros and cons to justify its judgment on image quality. 
\end{itemize}

\subsection{Dataset Construction}
\label{subsec:dataset_construction}

Data serves as the key factor for training MLLMs. 
We follow the scheme of supervised fine-tuning~\cite{llava, lamm} to train \model. 
Supervised fine-tuning requires collecting \{images, question, response\} data pairs, where ``images'' are the ones to be evaluated, ``question'' describes the task, and ``response'' is the reference answer.
In this section, we describe in detail our selection of images and the collection of questions and responses for the three tasks.

\textbf{Image collection} for the IQA dataset consists of two aspects, one is the selection of reference images, and the other is the collection of distorted images.
Existing works usually collect a large number of distorted images on a small number of reference images~\cite{tid2013, pieapp, gu2020image, gu2020pipal}. 
However, the semantic richness of the training images is also very important. 
In our work, we collect reference images and their corresponding distorted images from the BAPPS dataset~\cite{bapps}. 
BAPPS dataset contains 197k highly diverse samples, meeting the diversity requirements of \model training. 
For each sample, there is a reference image and its two distorted versions, as given in \cref{fig:dataset}b and \ref{fig:dataset}c. 
BAPPS dataset also provides human annotations, indicating which distorted image is more similar to the reference one, \textit{i.e.}, has better perceptual quality. 
These annotations can be used to build our dataset and validate our newly annotated data.

\textbf{Question collection}.
Users may express similar questions in different sentences, like the two questions in \cref{fig:teaser}. 
To encourage the robustness to users' questions, for each task, we first leverage GPT-4 to generate 30 questions.
We then manually remove ambiguous and duplicate ones and modify inaccurate ones to form a set of 10 questions (see \supp).
During training and testing, we randomly sample a question from the question set to construct the data pair.

\textbf{Response collection}.
A straightforward way to collect high-quality training texts is manually writing~\cite{llama2, instruct_gpt}.
However, when annotators are inexperienced or tired, human-written texts can lead to biases and uneven quality.
In this work, we use advanced LLMs to assist annotators in generating structured training texts, as shown in \cref{fig:dataset}.
We first collect the information that we want the texts to describe through a carefully designed questionnaire. 
Answering questions greatly reduces the possibility of ambiguity among annotators and ensures that the information is structured.
We then use GPT-4 to convert our annotated questionnaire results into natural language. 
Finally, the outputs of GPT-4 are modified and improved by the annotators to correct errors, eliminate ambiguities, and add important information. 
This process greatly reduces the difficulty of collecting training texts and improves the quality of the training texts.

Next, we introduce the details of the questionnaire for different tasks.

\textit{Task--1: Quality Description}.
A distorted image and its reference image are shown to annotators. 
\cref{fig:dataset}a shows our questionnaire with three parts: texture, distortion, and overall quality. 
For the texture part, annotators are asked to select the one that best matches the image from a list containing 11 typical texture types, including object edges, bricks, fabrics, plants or foliage, architectures, artificial strips, hairs or furs, faces, sky or clouds, stones or ground, and water surface. 
These 11 types are selected based on the existing IQA~\cite{gu2020pipal} and texture recognition~\cite{minc} research. 
Additionally, annotators are asked to indicate whether the texture is ``clearly identifiable'', ``barely identifiable'', or ``completely damaged''. 
Regarding the distortion part, we ask annotators to summarize with the following five aspects: ``brightness'', ``color'', ``noise'', ``artifacts'', and ``blurriness''. 
For each distortion, we use three levels for evaluation: ``undistorted'', ``mild'', and ``severe''. 
This can express most distortions that appear in images. 
Finally, annotators need to comment on the overall quality of the image into three levels: ``nearly undistorted'', ``mild distortion'', and ``severe distortion''.

\textit{Task--2: Quality Comparison}. 
BAPPS dataset already includes binary comparison labels (\textit{i.e.}, Image A or Image B is better) for all image pairs.
To convert these comparison labels into textural responses, we first build a response pool using GPT-4, including 20 generated sentences for ``Image A is better'' and another 20 for ``Image B is better''. 
Then, for each comparison label, we randomly sample one response from the pool, as depicted in \cref{fig:dataset}b.
However, the diversity of language output poses a challenge to evaluation. 
For the convenience of evaluation, inspired by LLaVA-1.5~\cite{llava1.5}, we randomly sample half of the questions and add the following short answer prompt: ``Answer the question using a single word or phrase''. 
Correspondingly, the response will be a single phrase like ``Image A'' or ``Image B'' indicating the less distorted image.

\textit{Task--3: Comparison Reasoning}. 
As shown in \cref{fig:dataset}c, annotators are given two distorted images and the reference image. 
The annotation pipeline is similar to Task--1.
Annotators compare two distorted images from the five kinds of distortions and the overall distortion using five options: ``superior'' ($>>$), ``slightly superior'' ($>$), ``roughly equal'' ($\approx$), ``slightly inferior'' ($<$), and `inferior'' ($<<$).

\begin{table}[t]
\centering
\scriptsize
\setlength\tabcolsep{5pt}
\caption{\textbf{Statistics} of our constructed \dataset dataset with respect to different tasks and dataset splits.}
\label{tab:dataset_number}
\begin{tabular}{c|ccc}
\toprule
& \# Task--1  & \# Task--2  & \# Task--3 \\
& Quality Description & Quality Comparison & Comparison Reasoning \\
\midrule
Training / Validation & 1,115 / 50 & 115,646 / 9,440 & 3,739 / 200 \\
\bottomrule
\end{tabular}
\end{table}

\textbf{Dataset statistics.} 
The statistics of our dataset are presented in \cref{tab:dataset_number} (more in \supp). 
Our dataset comprises 5,104 detailed high-quality samples (Task--1 and Task--3), along with 115,646 brief templated samples (training set of Task--2). 
The validation set of Task--2 is the same as the ``Traditional'' and ``CNN'' categories (two sets of distortions) in BAPPS's validation set~\cite{bapps}. 
Each training sample is individually annotated by one annotator. 
In the validation set, samples are annotated by two annotators only if they reach a consensus. 
As a means of verification, the annotated ``Overall Comparison'' judgments in Task--3 exhibit a quite high consistency rate of 84.3\% with the ground-truth judgments in the BAPPS dataset. 
For brevity, the three tasks will be shortened to \textit{description}, \textit{comparison}, and \textit{reasoning} in the following.

\section{\model Framework}\label{sec:method}

\subsection{Model Architecture}\label{subsec:model}

\begin{figure*}[tb]
    \centering
    \includegraphics[width=1.0\linewidth]{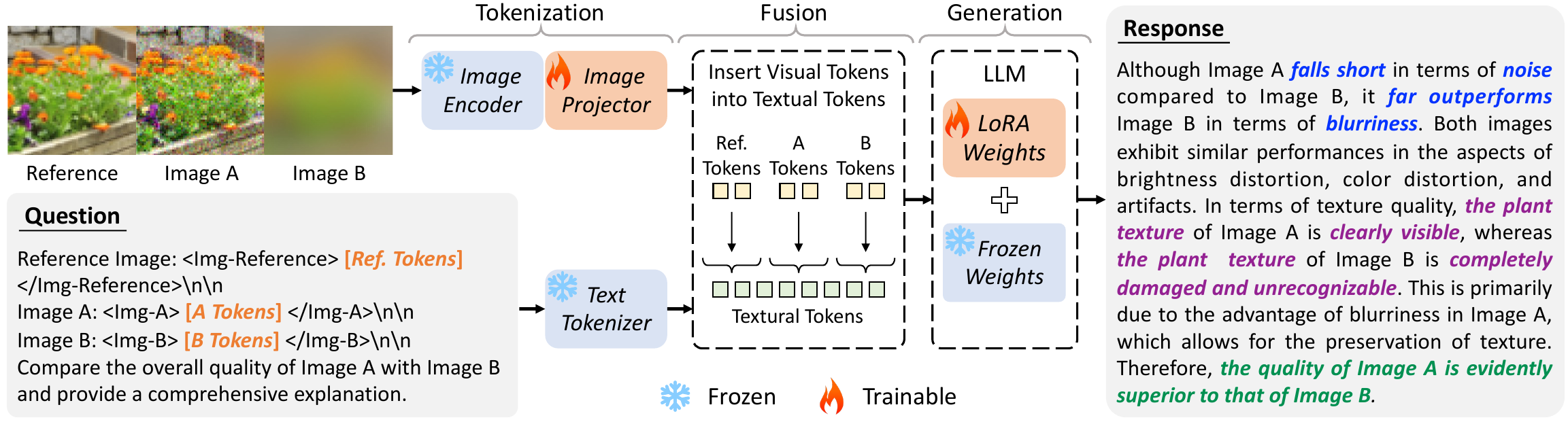}
    \caption{
        \textbf{Framework of \model}. 
        A frozen pre-trained image encoder is employed to encode images to visual tokens, followed by a trainable image projector to project visual tokens to textual space. 
        The question texts are tokenized by a text tokenizer. 
        Visual tokens and textual tokens are then fused and jointly processed by an LLM, fine-tuned through the LoRA technique~\cite{lora}. 
        Our model is capable of producing comprehensive and informative explanations for image quality comparisons. 
    }
    \label{fig:model}
\end{figure*}

\cref{fig:model} shows the workflow of our \model. 
\model takes images and a quality-related question as inputs, and generates a descriptive text as the response. 
In \textit{comparison} or \textit{reasoning} tasks, there are three input images: a reference image and Image A and B. 
In \textit{description} task, there are two input images: a reference image and a distorted image. 
The input images and the question are first tokenized, then fused, finally processed by the LLM for response generation.

\textbf{Tokenizing input images and question}. 
As shown in \cref{fig:model}, we employ a frozen CLIP pre-trained ViT-L/14~\cite{clip} as the image encoder to represent the input images as visual tokens. 
Then, the question texts are tokenized to textual tokens by SentencePiece tokenizer~\cite{sentencepiece}.
The visual tokens cannot be directly processed by the LLM due to different embedding spaces, so we use a trainable image projector to map visual tokens into the textual space as in~\cite{llava, minigpt4}.

\textbf{Token fusion}.
We insert the visual tokens into pre-defined positions within the textual tokens for token fusion. As show in \cref{fig:model}, [\textit{Ref. Tokens}], [\textit{A Tokens}], [\textit{B Tokens}] in the input question are these pre-defined positions.

\textbf{Response generation using LLM}.
The fused tokens are fed into LLM to generate the final response. 
Our \model is compatible with different LLMs (Vicuna-v0-7B~\cite{vicuna}, Vicuna-v1.5-7B~\cite{vicuna}, and LLaMA-2-chat-7B~\cite{llama2}), as shown in \cref{tab:ablation_llm}. 
Pre-trained LLMs do not work well on IQA tasks, and need to be fine-tuned on our dataset. 
However, complete LLM fine-tuning is resource-intensive and data-hungry, requiring tens of thousands of high-quality human-written texts~\cite{instruct_gpt}. 
To get around of data shortage issue, we resort to LoRA~\cite{lora}, an efficient LLM fine-tuning technique, which selectively adjusts only a small set of parameters in the LLM. 
Following~\cite{lora, lamm}, we apply LoRA to fine-tune the projection layers in all self-attention modules.

\textbf{Unique tag to distinguish multiple images}.
Existing MLLMs~\cite{llava, minigpt4} are primarily designed to handle a single input image. They insert the visual tokens between the start (\eg, \texttt{<Img>}) and end (\eg, \texttt{</Img>}) tags. 
A simple extension to multi-image input is using textual cues to distinguish images, \eg, adding ``Reference Image:'', ``Image A:'', and ``Image B:'' before visual tokens. 
However, this simple extension sometimes fails to distinguish images, probably because the proportion of these short textual cues in the full texts is too small. 
To mitigate this, motivated by~\cite{zhao2023mmicl}, we adopt the unique tag for each image. 
In \textit{comparison} and \textit{reasoning} tasks, we select \texttt{<Img-Reference>}, \texttt{<Img-A>}, and \texttt{<Img-B>} as start tags, adhering to the HTML rule by adding a trailing slash (``/'') in the end tags. 
In \textit{description} task, the reference image retains the same tags, while the distorted image employs the original tags, \texttt{<Img></Img>}.

\subsection{Training Scheme}

\textbf{Multi-source training data}. 
As stated in \cref{sec:introduction}, our training images come from three sets:
(1) 115,646 brief templated data of Task--2 (\textit{comparison}), 
(2) 4,854 high-quality data of Task--1 (\textit{description}) and Task--3 (\textit{reasoning}), which are duplicated by 20 times to increase the training weights, 
and (3) 48,734 content description data (Detailed Description dataset in~\cite{lamm}, duplicated 2 times during training), which are unrelated to IQA. 
The abundant templated data principally bridge images and descriptive texts in quality-related tasks. 
The limited yet high-quality data foster the model's descriptive and reasoning skills. 
The IQA-unrelated content description data can serve as regularization, given the limited text diversity of our IQA data for an MLLM. 
Experimental results in \cref{tab:joint_train} prove the effectiveness of the three sources of training data.

\textbf{Training objective}.
Following existing works~\cite{gpt3, lamm, llava}, the training objective of \model is the next token prediction loss: a cross-entropy loss between predicted and ground-truth next tokens. 
Only the tokens in the responses contribute to the loss computation. 
Also, only the image projection layer and LoRA parameters are optimized, comprising a mere 0.25\% of the total parameters (7B).

\section{Experiments}\label{sec:experiments}

This section discusses experimental setups and results. 
In LoRA, the rank and scale factor are both set as 16. 
In each attention layer of LLM, the projection weights of ``query'', ``key'', ``value'', and ``output'' are adjusted using two LoRA delta parameters. 
\model is trained for 1 epoch with batch size 64 on 8 GPUs (NVIDIA RTX A6000 48G). 
Adam optimizer with $(\beta_1, \beta_2) = (0.9, 0.95)$, weight decay 0.001, and learning rate $5e^{-4}$ is used. 
The training is completed in around 12 hours. 
See \supp for details and the training and inference costs.

\subsection{Metrics}

Unlike traditional score-based IQA methods, evaluating the diverse, descriptive, and textual results of multi-modal IQA methods is not trivial. Therefore, we adopt three different kinds of metrics for a comprehensive evaluation.

\textbf{Accuracy, SRCC, and PLCC}.
For \textit{comparison} task, we employ the accuracy metric. 
MLLMs usually produce diverse textual outputs, necessitating transformation to bi-classification results (\textit{i.e.}, Image A or Image B is better) for accuracy calculation. 
As described in the \textit{Task--2} part in \cref{subsec:dataset_construction}, we add the short answer prompt ``Answer the question using a single word or phrase'' during training, guiding \model to directly output bi-classification results. 
To further evaluate the alignment between the comparison results and human perception, we conduct pair-wise comparison and adopt voting method to convert the bi-classification results to quality scores. 
The quality scores are assessed using Spearman Rank Correlation Coefficient (SRCC) and Pearson Linear Correlation Coefficient (PLCC) following score-based IQA methods~\cite{LIQE, CLIP-IQA}.

\textbf{GPT-4 score}.
For \textit{description} and \textit{reasoning} tasks, following~\cite{vicuna, llava}, we utilize the GPT-4 score for assessment. 
Specifically, we provide GPT-4 with all information in the human-labeled questionnaire as context. 
Then, we give GPT-4 both the model-generated response and the corresponding ground truth response. 
Based on the context, GPT-4 evaluates the helpfulness, relevance, accuracy, and level of detail of these two responses, and gives an overall score on a scale of 0 to 10, where a higher score indicates better quality. 
Finally, the relative score with regard to the ground truth response is reported as the GPT-4 score.

\textbf{Reasonable rate by human evaluators}.
We observe that GPT-4 score exhibits excessive confidence in some low-quality responses where a wrong response even receives a GPT-4 score over 60\% (see \supp). 
Thus, for a comprehensive evaluation, given both the images and responses, human evaluators label each response as reasonable or not. 
A reasonable response should adhere to three criteria: indicating the major one distortion, no severe comparison mistakes, and self-consistency (see \supp). 
The reasonable rate serves as another metric.

\subsection{Comparison with Score-based IQA Methods}\label{subsec:compare_iqa}

To demonstrate the effectiveness of \model, we compare it with score-based IQA methods on the \textit{comparison} task (Task--2). 
We evaluate four traditional IQA methods including PSNR, SSIM~\cite{ssim}, VIF~\cite{vif}, and FSIM~\cite{fsim}, and four deep-learning-based IQA methods including DeepIQA~\cite{deepiqa}, PieAPP~\cite{pieapp}, LPIPS~\cite{bapps}, and DISTS~\cite{dists}. 
Here our base LLM is selected as LLaMA-2-chat-7B~\cite{llama2}.

\textbf{Quantitative results of quality comparison on BAPPS} are depicted in \cref{tab:bapps}a. 
Our \model surpasses the best traditional method, FSIM, by a large margin ($\sim$11\%). 
We also stably outperform the best deep-learning-based competitors, LPIPS and DISTS ($>$2.6\%), even near the human performance. 
We show in \cref{tab:ablation_llm} that the accuracy could be further enhanced with pre-training initialization (but with near $3\times$ training time). 
Unlike high-level perception tasks where multi-modal approaches usually lag behind single-modal methods~\cite{lamm}, we show that multi-modal IQA methods can surpass score-based counterparts in the quality comparison task. 
Nevertheless, the quantitative comparison is not the key issue, we pay more attention to the description and reasoning abilities.

\textbf{Quantitative results of quality comparison on multiple IQA datasets} are shown in \cref{tab:more_datasets}b. 
In this experiment, we include additional training datasets including PIPAL~\cite{gu2020pipal}, KADID~\cite{kadid}, and CSIQ~\cite{csiq}. 
These datasets consist of reference images and corresponding distorted versions. 
To adapt them to \textit{comparison} task, we reorganize the datasets by randomly selecting a reference image and its two distortions. 
The ground truth is established by comparing the MOS or DMOS of these two distortions. 
The questions and responses are constructed with pre-defined templates. 
During evaluation, besides these datasets' non-overlapped validation split, we also evaluate on unseen benchmarks including TID2013~\cite{tid2013}, LIVE~\cite{live}, LIVE-MD~\cite{livemd}, MDID2013~\cite{mdid}. 
We conduct pair-wise comparison and adopt voting method to transform the bi-classification results to quality scores. 
SRCC and PLCC metrics are reported. 
\model stably outperforms two baselines, validating its effectiveness across multiple datasets.

\begin{table}[t]
\caption{\textbf{Quantitative results of quality comparison task} on traditional IQA benchmarks. 
Our \model achieves the state-of-the-art performance. 
}
\begin{minipage}[t]{0.44\textwidth}
\setlength\tabcolsep{5pt}
\scriptsize
(a) Accuracy in Traditional / CNN distortion categories on BAPPS benchmark~\cite{bapps}. 
\begin{flushleft}
\begin{tabular}{c|c|c}
\toprule
Type & Method & Comparison \\
\midrule
Oracle & Human & 80.8 / 84.4 \\
\midrule
\multirow{4}{*}{\makecell[c]{Tradi-\\tional}}
& PSNR & 59.9 / 77.8 \\
& SSIM~\cite{ssim} & 60.3 / 79.1  \\
& VIF~\cite{vif} & 55.6 / 74.4  \\
& FSIM~\cite{fsim} & 62.7 / 79.4  \\
\midrule
\multirow{4}{*}{\makecell[c]{Learning}}
& DeepIQA~\cite{deepiqa} & 70.3 / 79.4  \\
& PieAPP~\cite{pieapp} & 72.7 / 77.0  \\
& LPIPS~\cite{bapps} & 76.0 / 82.8  \\
& DISTS~\cite{dists} & 77.2 / 82.2  \\
\midrule
\multicolumn{2}{c|}{\model (Ours)} & \textbf{80.3} / \textbf{84.2} \\
\bottomrule
\end{tabular}
\label{tab:bapps}
\end{flushleft}
\end{minipage}
\hfill
\begin{minipage}[t]{0.53\textwidth}
\setlength\tabcolsep{2pt}
\renewcommand{\arraystretch}{1.32}
\scriptsize
(b) SRCC / PLCC metrics on benchmark datasets including KADID~\cite{kadid}, CSIQ~\cite{csiq}, TID2013~\cite{tid2013}, LIVE~\cite{live}, LIVE-MD~\cite{livemd}, MDID2013~\cite{mdid}. 
\begin{flushright}
\begin{tabular}{c|cc|c}
\toprule
\multirow{3}{*}{Datasets} & \multicolumn{3}{c}{Methods} \\ & \multirow{2}{*}{FSIM~\cite{fsim}} & \multirow{2}{*}{LPIPS~\cite{bapps}} & \multirow{2}{*}{\makecell[c]{DepictQA\\(Ours)}} \\
& & & \\
\midrule
KADID & 0.855/0.857 & 0.799/0.803 & \textbf{0.939}/\textbf{0.944} \\
CSIQ  & 0.937/0.937 & 0.905/0.926 & \textbf{0.940}/\textbf{0.954} \\
TID2013 & 0.841/0.875 & 0.798/0.851 & \textbf{0.853}/\textbf{0.886} \\
LIVE & 0.894/0.908 & 0.906/\textbf{0.921} & \textbf{0.913}/0.914 \\
LIVE-MD & 0.877/0.910 & 0.897/0.913 & \textbf{0.905}/\textbf{0.928} \\
MDID2013 & 0.717/0.770 & 0.731/0.764 & \textbf{0.820}/\textbf{0.835} \\
\bottomrule
\end{tabular}
\label{tab:more_datasets}
\end{flushright}
\end{minipage}
\end{table}

\subsection{Comparison with General Multi-modal LLMs}\label{subsec:compare_mllm}

We also compare \model with general MLLMs on \textit{description} (Task--1) and \textit{reasoning} (Task--3) tasks. 
We also transform the reasoning responses to bi-classification results to calculate the comparison accuracy. 
We choose four MLLMs as baselines: LLaVA~\cite{llava}, LLaVA-1.5~\cite{llava1.5}, MiniGPT4~\cite{minigpt4}, and LAMM~\cite{lamm}. 
We provide explicit instructions to inform the MLLMs of the task definition.

\textbf{Quantitative results} are illustrated in \cref{tab:compare_mllm}. 
LLaVA and MiniGPT4 (LLaMA-2-chat) fail in IQA tasks, yielding either nearly identical results or irrelevant results across most samples. 
Other general MLLMs exhibit poor performance, indicating their inadequacy for IQA tasks. 
However, after fine-tuning on our \dataset dataset, \model achieves significantly improved performance.
One possible reason for general MLLMs' inadequacy is that they are trained on single images, while our tasks require multi-image input. 
Therefore, we also test LLaVA-1.5 on \textit{description} task without a reference (\ie, single-image input). 
We then employ GPT-4 to verify whether ground-truth distortions are mentioned in the responses. 
The results (24\%) indicate some improvements over the full-reference setting (18\%). 
However, the overall performance remains sub-optimal. 
Qualitative results in \supp show the unsatisfying results of general MLLMs.

\textbf{Qualitative results}. 
Three qualitative results of \textit{reasoning} task are depicted in \cref{fig:teaser} and \cref{fig:model}. 
More qualitative results and failure cases of \textit{description}, \textit{comparison}, and \textit{reasoning} tasks are illustrated in \supp.

\begin{table}[t]
\centering
\scriptsize
\setlength\tabcolsep{2pt}
\caption{
\textbf{Comparison with general MLLMs}. 
All LLMs have 7B parameters. ``LLaMA-2c'' means LLaMA-2-chat. 
Metric for comparison: accuracy. 
Since general MLLMs can produce responses without explicit comparison results, accuracy is reported with these responses included / excluded. 
Metric for description and reasoning: reasonable rate / GPT-4 score. 
General MLLMs are not capable of IQA tasks. 
}
\label{tab:compare_mllm}
\begin{tabular}{c|cccccc|c}
\toprule
Method & \makecell[c]{LLaVA\\\cite{llava}} & \makecell[c]{LLaVA-1.5\\\cite{llava1.5}} & \makecell[c]{MiniGPT4\\\cite{minigpt4}} & \makecell[c]{MiniGPT4\\\cite{minigpt4}} & \makecell[c]{LAMM\\\cite{lamm}} & \makecell[c]{LAMM\\\cite{lamm}} & \makecell[c]{\model\\(Ours)} \\
\midrule
LLM (7B) & LLaMA-2c & Vicuna-v1.5 & Vicuna-v0 & LLaMA-2c & Vicuna-v0 & LLaMA-2c & Vicuna-v1.5 \\
\midrule
Description & fail & 18.0 / 65.8 & 16.0 / 49.9 & fail & 12.0 / 62.5 & 8.0 / 57.4 & \textbf{64.0} / \textbf{76.2} \\
Comparison & fail & 43.0 / 50.6 & 38.0 / 46.3 & fail & 53.0 / 55.8 & 44.0 / 48.9 & \textbf{82.0} \\
Reasoning & fail & 7.0 / 63.9 &  1.0 / 42.4 & fail & 4.0 / 58.1 & 5.0 / 52.4 & \textbf{53.0} / \textbf{76.4} \\
\bottomrule
\end{tabular}
\end{table}

\subsection{Ablation Studies}

We conduct extensive ablation studies to verify the effectiveness of our methodologies. 
The LLM used in ablation studies is Vicuna-v1.5-7B~\cite{vicuna} if not specified.

\textbf{Effects of multi-source training data} are detailed in \cref{tab:joint_train}. 
(1) Task-2 data (\textit{comparison}) remarkably improves comparison accuracy (\#2 \vs\#4). 
Though these texts are brief and templated, the abundant samples still help bridge the images and texts in quality-related tasks. 
(2) As evidenced by \#1 \vs\#4 and \#3 \vs\#5, Task-3 data (\textit{reasoning}) is necessary for a robust reasoning ability. 
Task-1 data (\textit{description}) boosts reasoning metric (\#4 \vs\#5), mainly by helping identify distortions. 
Therefore, the high-quality texts of Task--1 and Task--3 are necessary to depict the image quality with language. 
(3) However, as shown by \#1 \vs\#3 and \#4 \vs\#5, Task-1 data (\textit{description}) harms the comparison performance because of the gap between the two tasks in prompts and the number of input images. 
(4) Fortunately, these negative effects can be eliminated by content description data (\#5 \vs\#6), which serves as an regularization in light of \dataset's limited text variety for a huge MLLM. 
Additionally, the content description data helps enrich the text diversity, as shown in \supp.

\begin{table}[t]
\begin{minipage}[t]{0.5\textwidth}
\setlength\tabcolsep{1pt}
\renewcommand{\arraystretch}{1.13}
\begin{flushleft}
\scriptsize
\caption{\textbf{Ablation studies of multi-source training data}. 
``Cont.'' means the content description data. Metric for comparison: accuracy within Traditional / CNN categories. Metric for reasoning: reasonable rate / GPT-4 score. 
Multi-source training data stably improves the performance. 
}
\begin{tabular}{ccccc|cc}
\toprule
\multirow{2}{*}{\#} & \multicolumn{4}{c|}{Training Data} & \multirow{2}{*}{Comparison} & \multirow{2}{*}{Reasoning} \\
 & Task2 & Task1 & Task3 & Cont. &  &  \\
\midrule
1 & \ding{51} & & & & 78.6/82.5 & N/A \\
2 & & & \ding{51} & & 66.4/65.4 & 31.0/74.2  \\
3 & \ding{51} & \ding{51} & & & 76.3/80.5 & N/A \\
4 & \ding{51} & & \ding{51} & & 79.7/83.3 & 41.0/74.3 \\
5 & \ding{51} & \ding{51} & \ding{51} & & 78.1/82.9 & 45.0/\textbf{77.2} \\
6 & \ding{51} & \ding{51} & \ding{51} & \ding{51} & \textbf{80.0}/\textbf{83.8} &  \textbf{53.0}/76.4 \\
\bottomrule
\end{tabular}
\label{tab:joint_train}
\end{flushleft}
\end{minipage}
\hfill
\begin{minipage}[t]{0.47\textwidth}
\setlength\tabcolsep{1pt}
\begin{flushright}
\scriptsize
\caption{
\textbf{Ablation studies of LLMs and initialization}. 
``High-level'' means the MLLM is pre-trained on high-level perception tasks. 
Metric for comparison: accuracy within Traditional / CNN categories. 
\model is compatible with different LLMs. 
Pre-training on high-level perceptual tasks brings improvement. 
}
\begin{tabular}{ccc|c}
\toprule
\# & LLM & Init. & Comparison  \\
\midrule
1 & Vicuna-v0-7B & Random & 79.5/83.7 \\
2 & Vicuna-v0-7B & High-level & \textbf{79.8}/\textbf{84.2} \\
\midrule
3 & Vicuna-v1.5-7B & Random & 80.0/83.8 \\
4 & Vicuna-v1.5-7B & High-level & \textbf{81.2}/\textbf{85.3} \\
\midrule
5 & LLaMA-2-chat-7B & Random & \textbf{80.3}/84.2 \\
6 & LLaMA-2-chat-7B & High-level & 80.1/\textbf{85.1} \\
\bottomrule
\end{tabular}
\label{tab:ablation_llm}
\end{flushright}
\end{minipage}
\end{table}

\begin{figure}[t]
\begin{minipage}[t]{0.5\textwidth}
\begin{flushleft}
\includegraphics[width=1.0\linewidth]{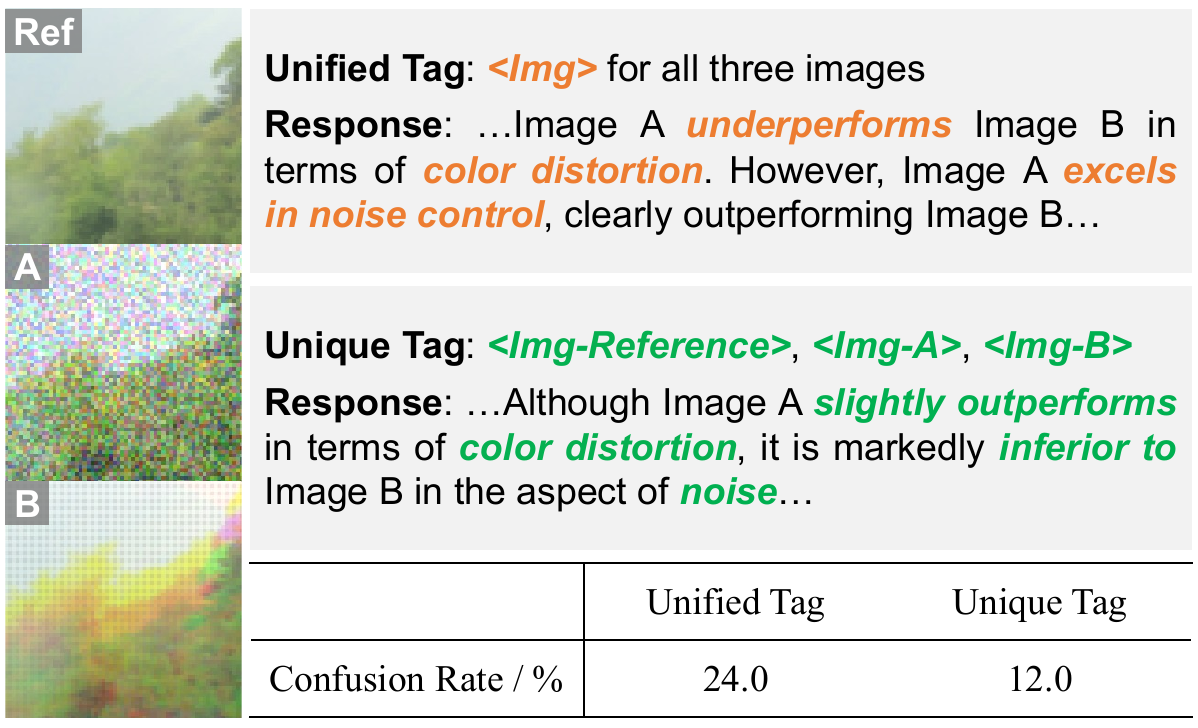}
\caption{
    \textbf{Unique tag alleviates the confusion problem} using clearer instructions. The confusion rate drops dramatically. 
}
\label{fig:multi_tags}
\end{flushleft}
\end{minipage}
\hfill
\begin{minipage}[t]{0.48\textwidth}
\begin{flushright}
\scriptsize
\includegraphics[width=1.0\linewidth]{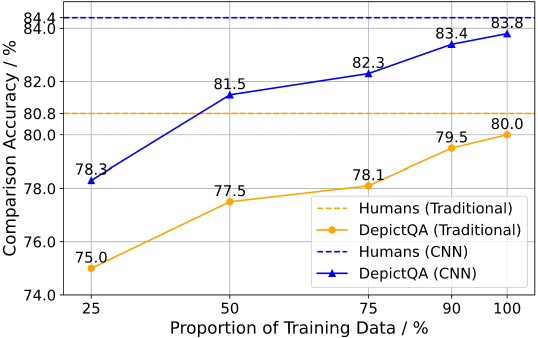}
\caption{
    \textbf{The comparison performance gradually increases} with the size of training data increasing. 
}
\label{fig:ablation_datasize}
\end{flushright}
\end{minipage}
\end{figure}

\textbf{Unique tag} effectively mitigates the confusion problem, detailed in \cref{fig:multi_tags}. 
Confusion occurs when distortions in one image are mistakenly attributed to another.
To quantify this, we manually review 50 responses and compute the confusion rate.
With the unified tag, the model needs to distinguish images through textual hinds, as stated in \cref{subsec:model}, leading to a 24\% confusion rate. 
One unique tag for each image significantly reduces the confusion rate to 12\%. 
Two alternative methods to distinguish multiple images (\ie, image embedding, unique projector) are studied in \supp, showing the advantage of unique tag.

\textbf{The size of training data} is studied in \cref{fig:ablation_datasize}. 
As the size of training data increases, the comparison performance gradually increases. 
Thus the quantity of the training data still plays a key role in the MLLM-based method.

\textbf{LLMs and the initialization methods} are studied in \cref{tab:ablation_llm}. 
For high-level initialization, trainable parameters are pre-trained on high-level perception tasks (\textit{e.g.}, description and captioning). 
(1) Overall, our \model is compatible with different LLMs, stably yielding high performance. 
Also, the performance is slightly improved with more advanced LLMs (\eg, Vicuna-v1.5 and LLaMA-2-chat). 
(2) With the same LLM, pre-training on high-level perceptual tasks further enhances the overall performance.
Nevertheless, pre-training on high-level perceptual tasks will also increase the training time by nearly three times.

\subsection{Extensions}\label{subsec:extensions}

\begin{table}[t]
\begin{minipage}[t]{0.45\textwidth}
\begin{flushleft}
\centering
\setlength\tabcolsep{17pt}
\scriptsize
\captionof{table}{
    \textbf{Full-reference / non-reference performance}. Metric for comparison: accuracy. Metric for description and reasoning: GPT-4 score. 
}
\label{tab:non_refer}
\begin{tabular}{c|c}
\toprule
Task & Performance \\
\midrule
Description & 76.4 / 84.5 \\
Comparison  & 82.3 / 78.5 \\
Reasoning   & 77.2 / 78.8 \\
\bottomrule
\end{tabular}
\end{flushleft}
\end{minipage}
\hfill
\begin{minipage}[t]{0.50\textwidth}
\renewcommand{\arraystretch}{1.125}
\begin{flushright}
\centering
\setlength\tabcolsep{4pt}
\scriptsize
\caption{\textbf{Comparison with the proprietary GPT-4V~\cite{gpt4v}}. Metric for comparison: accuracy. Metric for reasoning: reasonable rate / GPT-4 score. 
}
\label{tab:gpt4v}
\begin{tabular}{c|cc}
\toprule
Method & Comparison & Reasoning \\
\midrule
GPT-4V~\cite{gpt4v} & 65.0 & \textbf{52.0} / \textbf{106.4} \\
\midrule
\model (ours) & \textbf{90.0} & \textbf{53.0} / 77.6 \\
\bottomrule
\end{tabular}
\end{flushright}
\end{minipage}
\end{table}

\textbf{Non-reference applications}. 
Although our dataset is initially gathered in a full-reference setting, we demonstrate its utility in non-reference applications. 
We generate non-reference training data by simply removing the reference image in our original dataset. 
This non-reference data is combined to co-trained our model. 
The results are shown in \cref{tab:non_refer}. 
For the \textit{description} task, non-reference performance stably surpasses full-reference, possibly because single-image tasks are easier than multi-image tasks. 
Our model keeps its high performance in the non-reference setting, proving the feasibility of the non-reference extension.

\textbf{Comparison with GPT-4V}~\cite{gpt4v} is conducted on the first 100 samples in the validation set of Task--3 (\textit{reasoning})\footnote[1]{The results were tested on the preview web version released in Nov. 2023. A comprehensive evaluation of the stable API version, \texttt{gpt-4-turbo}, is given in~\cite{depict_v2}.}. 
We request GPT-4V to complete both \textit{comparison} and \textit{reasoning} tasks. 
GPT-4V is given detailed instructions and two examples for task description. 
As shown in \cref{tab:gpt4v}, \model and GPT-4V have mutual advantages and disadvantages. 
The GPT-4 score of GPT-4V is quite high, even higher than human-annotated responses ($>$100\%), because of the linguistic fluency and detailed description of contents. 
However, we achieve comparable performance at the human-evaluated reasonable rate. 
Moreover, \model is significantly better than GPT-4V in quality comparison. 
Note that GPT-4V is close-source and expensive to access, thus developing an effective MLLM for IQA is worthwhile. 
See \supp for qualitative results and failure cases of GPT-4V.

\section{Conclusions and Limitations}\label{sec:conclusion}

In this preliminary attempt, we propose an MLLM-based IQA model, \model, demonstrating the possibility of depicting image quality with languages. 
There is still a long way to go for real-world application. 
(1) The amount and coverage of data are not sufficient, limiting the generalization. 
(2) The performance in \textit{description} and \textit{reasoning} tasks is not satisfying. 
(3) The distortion types can be more than five, even not pre-defined. Also, fine-grained comparisons on local details are preferred. 
(4) The voting method with pair-wise comparison is time-consuming. 
(5) Finally, whether MLLM-based IQA methods can take the place of score-based ones is still an open question. 
These belong to our future works.

\noindent \textbf{Acknowledgement}.
This work was sponsored by CUHK Direct Grants (RCFUS) No. 4055189, National Natural Science Foundation of China (Grant No. 62276251), and the Joint Lab of CAS-HK.

{
    \small
    \bibliographystyle{splncs04}
    \bibliography{ref}
}

\clearpage

\renewcommand\thesection{A\arabic{section}}
\renewcommand\thefigure{A\arabic{figure}}
\renewcommand\thetable{A\arabic{table}}  
\renewcommand\theequation{A\arabic{equation}}
\setcounter{section}{0}
\setcounter{equation}{0}
\setcounter{table}{0}
\setcounter{figure}{0}

\section*{Appendix}
\section{Overview}\label{appendix:sec:overview}

This \supp is structured as follows.
Dataset details  are described in \cref{appendix:sec:dataset}, followed by the methodology details in \cref{appendix:sec:details}. 
\cref{appendix:sec:ablation} provides additional ablation studies.
More quantitative results, qualitative results, and failure cases are presented in \cref{appendix:sec:results}.

\section{Dataset Details}\label{appendix:sec:dataset}

In this section, we provide additional descriptions and statistics of the introduced \dataset dataset.

\textbf{In-context learning for response generation}.
As detailed in the main paper, GPT-4~\cite{gpt4} is employed to convert human-annotated questionnaire options into a natural language paragraph. 
Yet, GPT-4 tends to produce responses that are overly verbose and potentially misleading. 
To mitigate this, we supply GPT-4 with two human-written examples to align the generated responses with a consistent writing style. 
Empirically, in-context learning effectively resolves these issues, yielding high-quality responses.

\textbf{Questions in the question set}. 
We show various question examples in the qualitative results of Task--1 (\textit{quality description}, \cref{appendix:fig:res_description_0} \& \ref{appendix:fig:res_description_1}), 
Task--2 (\textit{quality comparison}, \cref{appendix:fig:res_comparison_0} \& \ref{appendix:fig:res_comparison_1}), 
and Task--3 (\textit{comparison reasoning}, \cref{appendix:fig:res_reasoning_0}, \ref{appendix:fig:res_reasoning_1}, \& \ref{appendix:fig:res_reasoning_2}). 
This corroborates that users may express similar questions in different sentences, as discussed in the main paper.

\textbf{Statistics of the response length} in our \dataset dataset are illustrated in \cref{appendix:tab:dataset_number}. 
We present metrics on both word count and string length.
For the evaluation of Task--2 (\textit{quality comparison}), only the bi-classification ground-truth is utilized, so there is no need for statistics of the validation set.
The distribution of word length in our \dataset dataset is depicted in \cref{appendix:fig:wordlength}.

\begin{table}[t]
\centering
\scriptsize
\caption{
\textbf{Response length statistics} in our \dataset dataset, reported as word count / string length.
$\dag$ For Task--2's evaluation, only the bi-classification ground-truth is utilized, without the necessity for validation set statistics.
}
\setlength\tabcolsep{8pt}
\label{appendix:tab:dataset_number}
\begin{tabular}{c|ccc}
\toprule
 & Task--1 & Task--2 & Task--3 \\
 & Quality Description & Quality Comparison & Comparison Reasoning \\
\midrule
Training Set     & 135.1 / 888.9 & 9.4 / 52.2 & 143.3 / 886.4 \\
Validation Set & 143.8 / 947.3 & -$^\dag$     & 140.0 / 883.3 \\
\bottomrule
\end{tabular}
\end{table}

\begin{figure}[t]
\begin{minipage}[t]{0.42\textwidth}
\begin{flushleft}
\includegraphics[width=1.0\linewidth]{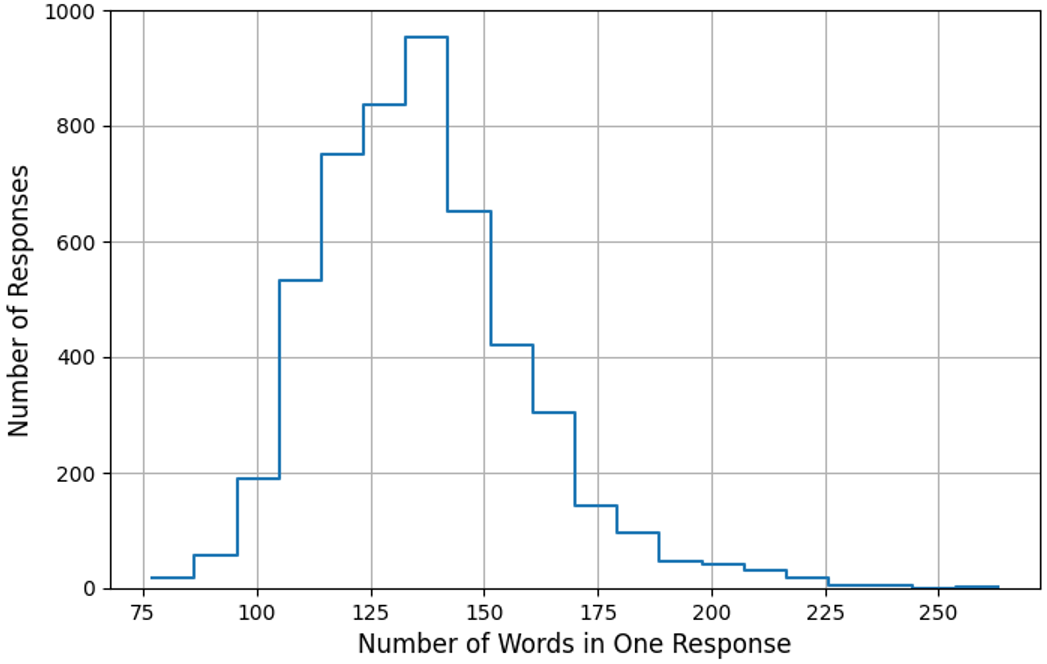}
\caption{
    \textbf{Distribution statistics of the word length} in our introduced \dataset dataset. 
}
\label{appendix:fig:wordlength}
\end{flushleft}
\end{minipage}
\hfill
\begin{minipage}[t]{0.55\textwidth}
\begin{flushright}
\scriptsize
\includegraphics[width=1.0\linewidth]{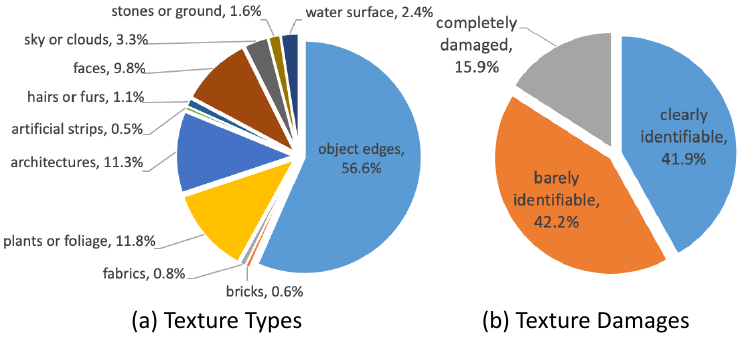}
\caption{
    \textbf{Statistics of texture-related options} in the questionnaire with respect to \textbf{(a)} texture types and \textbf{(b)} texture damages. 
}
\label{appendix:fig:texture}
\end{flushright}
\end{minipage}
\end{figure}

\textbf{Statistics of questionnaire options} are presented in \cref{appendix:fig:texture} and \cref{appendix:fig:distortion}. 
As detailed in the main paper, when annotating, annotators first complete a carefully designed questionnaire. 
Here we provide statistical breakdowns of the questionnaire's options. 
Texture-related option statistics are outlined in \cref{appendix:fig:texture}. 
Distortion-related option statistics appear in \cref{appendix:fig:distortion}. 
Due to the distinct distortion options in Task--1 (\textit{quality description}) and  Task--3 (\textit{comparison reasoning}), their counts are presented separately.

\begin{figure*}[t]
    \centering
    \includegraphics[width=1.0\linewidth]{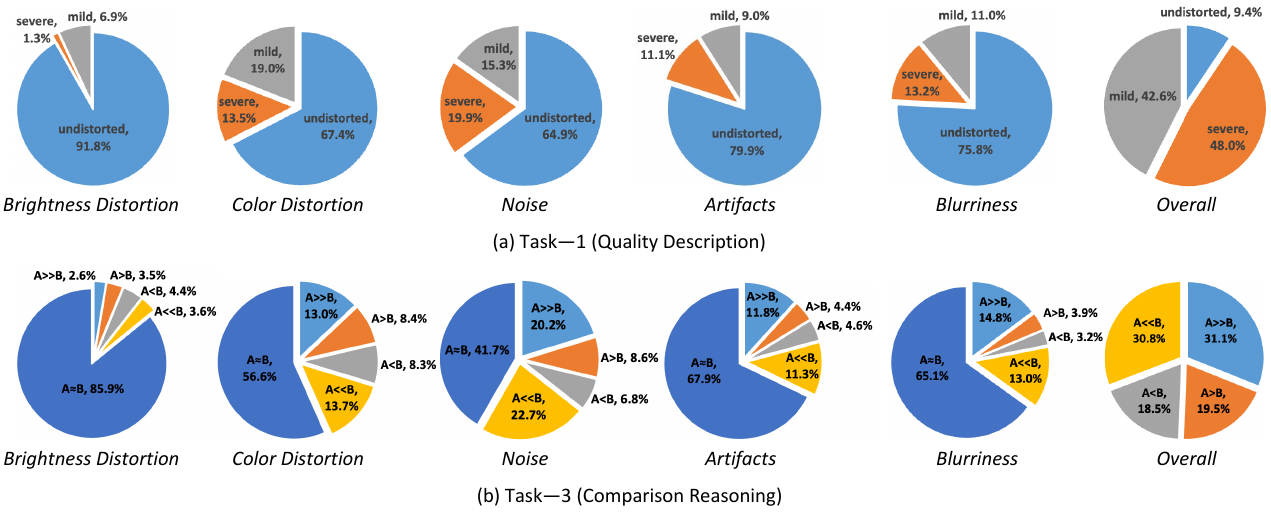}
    \caption{
        \textbf{Statistics of distortion-related options} in the questionnaire with respect to \textbf{(a)} Task--1 (\textit{quality description}) and \textbf{(b)} Task--3 (\textit{comparison reasoning}). 
        Note that there is no questionnaire for Task--2 (\textit{quality comparison}), hence no statistic is required. 
    }
    \label{appendix:fig:distortion}
\end{figure*}

\begin{figure}[t]
    \centering
    \includegraphics[width=0.95\linewidth]{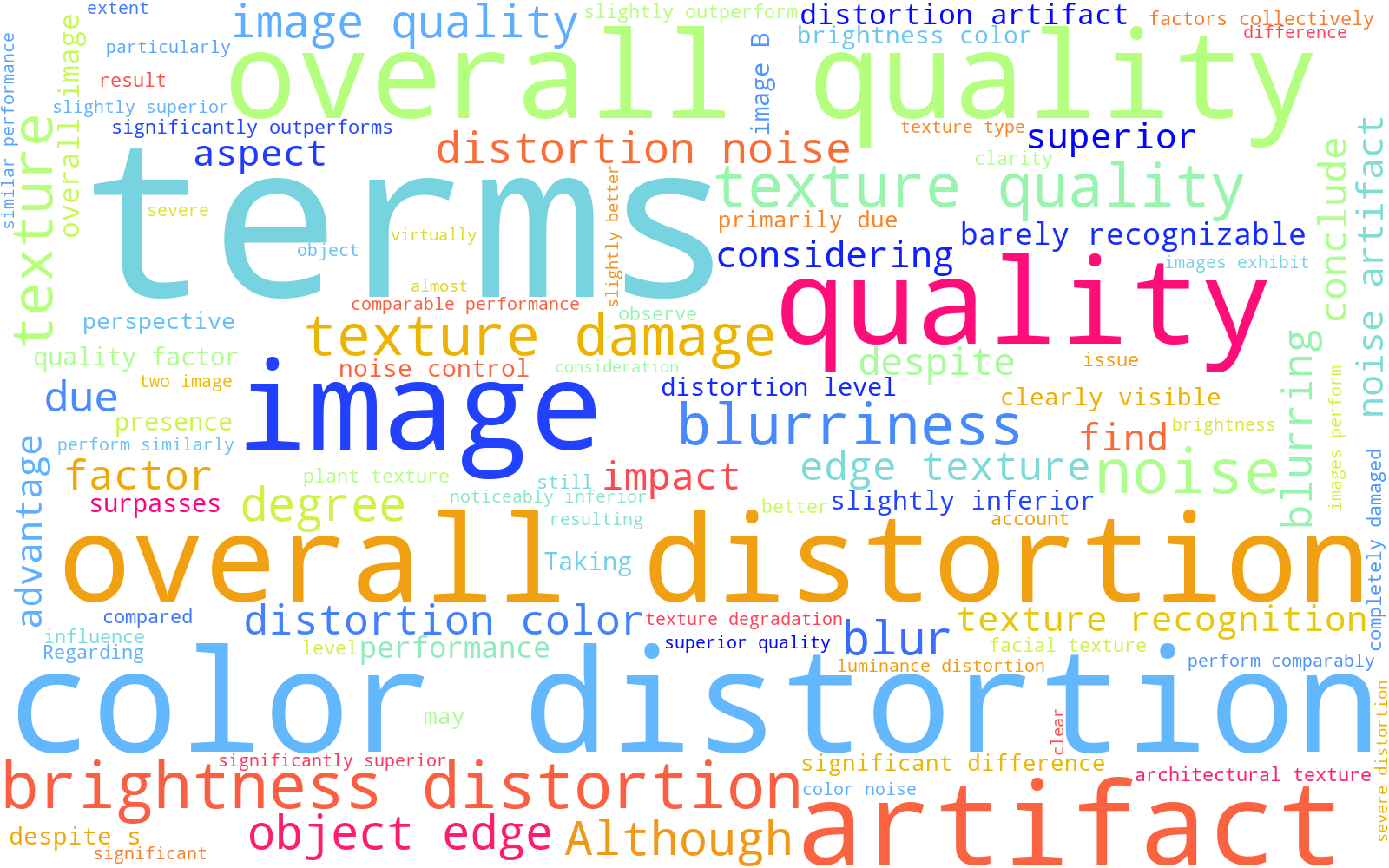}
    \caption{
        \textbf{Wordcloud} of \dataset dataset. The most frequent words in our \dataset dataset are all highly relevant to the visual quality of images. 
    }
    \label{appendix:fig:wordcloud}
\end{figure}

\textbf{Wordcloud} of our introduced \dataset dataset is depicted in \cref{appendix:fig:wordcloud}. 
Note that we manually exclude ``Image A'' and ``Image B'' while plotting \cref{appendix:fig:wordcloud}, as they are constant proper nouns across all texts. 
The most frequent words in our \dataset dataset (\eg, ``quality'', ``overall distortion'', ``color distortion'', and ``artifact'') are all highly relevant to the visual quality of images.

\section{Methodology Details}\label{appendix:sec:details}

In this section, we detail the model architecture, the calculation of metrics, and complexity and efficiency of our \model.

\subsection{Architecture}

\textbf{Model setup}. 
The image size in the BAPPS dataset~\cite{bapps} is $64 \times 64$. 
For other datasets including the detailed description dataset in LAMM~\cite{lamm} and existing IQA datasets, we keep the ratio between height and width and resize the short edge to 224. 
We then pad the image size to be a multiple of 14 with zero-padding, and partition the image into patches, with each sized at $14 \times 14$. 
We encode the image patches into visual tokens using the CLIP pre-trained ViT-L/14~\cite{clip}, with each token having the channel of 1024. 
To accommodate varying image resolutions, we interpolate (in bicubic mode) the position embedding in CLIP. 
The image projection layer projects these visual tokens to the hidden dimension of the LLM (\eg, Vicuna-v0-7B~\cite{vicuna}, Vicuna-v1.5-7B~\cite{vicuna}, and LLaMA-2-chat-7B~\cite{llama2}), which is 4096. 
In LoRA, the rank and scale factor are both set as 16. 
There are 32 attention layers in the LLM in total. 
In each attention layer, the projection weights of ``query'', ``key'', ``value'', and ``output'' are adjusted using two delta parameters with the shape of $4096 \times 16$ and $16 \times 4096$, respectively.

\subsection{Metrics}

\textbf{Accuracy, SRCC, and PLCC}. 
\model could directly output binary classification results (\textit{i.e.}, Image A or Image B is better) using the short answer prompt for accuracy computation. 
For other MLLMs, we deploy GPT-3.5 to convert varied textual comparison results into binary classification results for accuracy computation. 
Upon gathering approximately 10K textual comparison results and their binary classifications, we train a BERT~\cite{bert} model, achieving 97.5\% agreement with GPT-3.5 in the test dataset. 
To further evaluate the performance, we conduct pair-wise comparison and adopt voting method to convert the bi-classification results to quality scores. 
Following score-based IQA methods~\cite{LIQE, CLIP-IQA}, the quality scores are assessed using Spearman Rank Correlation Coefficient (SRCC) and Pearson Linear Correlation Coefficient (PLCC).

\textbf{GPT-4 score}. 
To compute the GPT-4 score for one response, we input GPT-4 with the following contents: ``context'', ``question'', ``answer\_1'', ``answer\_2'', and ``instruction''. 
The ``instruction'' directs GPT-4 to assess two answers (``answer\_1'' and ``answer\_2'') to the same ``question'' using the ``context'' knowledge, and then rate each answer on a scale of $[0, 10]$. 
The ``context'' contains all information in the human-labeled questionnaire, which is the correct knowledge for GPT-4 to evaluate answers. 
The ``question'' refers to the quality-related question posed to the MLLMs.
The ``answer\_1'' is actually the ground-truth response, which is labeled by human-evaluators. 
The ``answer\_2'' is the response of the MLLM to be evaluated. 
Then, the score of ``answer\_2'' (\ie, the MLLM to be evaluated) relevant to ``answer\_1'' (\ie, the ground-truth) serves as the GPT-4 score of ``answer\_2''. 
We calculate the final metric by averaging the scores across all samples. 
As discussed in the main paper, the GPT-4 score tends to be overly generous for low-quality responses, resulting in inflated scores. 
One example is shown in \cref{appendix:fig:gpt4_score}, where a fully wrong response receives a GPT-4 score of 66.7\%.

\begin{figure*}[t]
    \centering
    \includegraphics[width=1.0\linewidth]{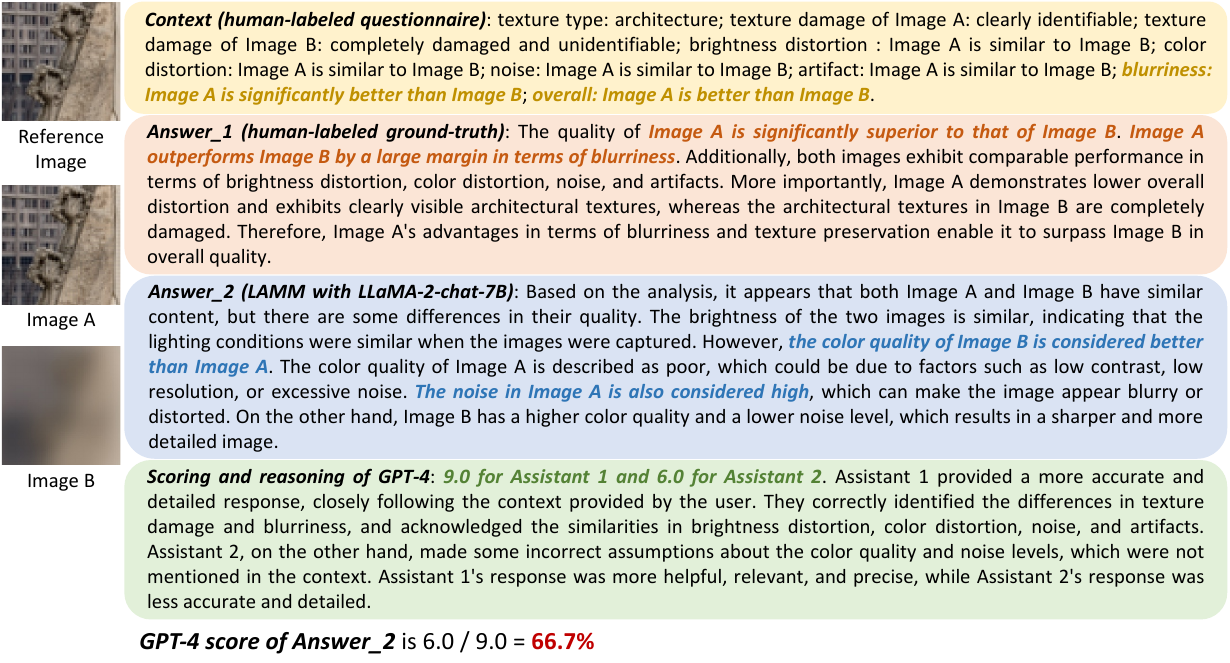}
    \caption{
        \textbf{GPT-4 score sometimes exhibits excessive confidence in low-quality responses}, though widely adopted as metrics in general MLLMs~\cite{vicuna, llava}. 
        In this case, a fully wrong response still receives a GPT-4 score of 66.7\%. 
    }
    \label{appendix:fig:gpt4_score}
\end{figure*}

\textbf{Reasonable rate evaluated by humans}. 
Human evaluators are requested to label each response as either reasonable or unreasonable, based on the provided images and corresponding responses. 
One ``reasonable'' response should adhere to three criteria. 
(1) The response must accurately identify the most important one quality issue that directly influences the comparison result. 
(2) The response should avoid seriously wrong comparison  results conflicting with human evaluation. 
(3) The response's description, reasoning, and conclusion should align closely. 
Minor errors not violating these three rules are permissible.

\subsection{Complexity and Efficiency}

\textbf{Training cost}. The trainable parameters in \model are 16M, constituting only 0.25\% of the total parameters (7B). \model is trained on 8 GPUs (NVIDIA RTX A6000 48G). The training is completed in around 12 hours.

\textbf{Inference cost}. The inference time depends on the response length and is tested on a single A6000 GPU on the BAPPS dataset~\cite{bapps}. 
(1) For \textit{comparison} task with short answer prompt (2 words), the inference time stands at approximately 2.20s / batch=32, transformed to 0.07s / sample. 
(2) For \textit{comparison} task without short answer prompt ($\sim$10 words), the inference time is 2.76s / batch=32 (\ie, 0.09s / sample) on average. 
(3) Regarding \textit{reasoning} task ($\sim$143 words), the inference time is around 20.86s / batch=32 (\ie, 0.65s / sample). 
Despite the incorporation of MLLMs that are known to be resource-intensive, our \model remains deployable on a single consumer GPU (\textit{e.g.}, RTX3090).

\section{Ablation Studies}\label{appendix:sec:ablation}

In this section, we conduct more ablation studies to verify the effectiveness of our methodologies.

\begin{table}[t]
\setlength\tabcolsep{10pt}
\centering
\scriptsize
\caption{
        \textbf{Comparison of global feature and local feature map}. 
        Metric for comparison: accuracy within Traditional / CNN categories. 
        Metric for reasoning: reasonable rate. 
        Local feature map surpasses global feature obviously. 
}
\label{appendix:tab:global_local}
\begin{tabular}{c|cc|cc}
\toprule
& \multicolumn{2}{c|}{Global Feature} & \multicolumn{2}{c}{Local Feature Map} \\
\midrule
& Comparison  & Reasoning & Comparison  & Reasoning \\
\midrule
Performance & 76.7 / 80.8 &  34.0 & \textbf{80.0} / \textbf{83.8} & \textbf{53.0} \\
\bottomrule
\end{tabular}
\end{table}

\textbf{Comparison of global feature and local feature map}. 
The ViT-L/14 image encoder produces a singular global feature token and a multi-token local feature map. 
Either can function as visual tokens of the MLLM. 
We compare the performance between the global feature and local feature map in \cref{appendix:tab:global_local}.
The global feature demonstrates lesser effectiveness than the local feature map, particularly in reasoning tasks. 
This disparity might come from the global feature's singular visual token, which restricts the amount of information conveyed.

\begin{table}[t]
\caption{\textbf{Ablation studies}. (a) Ablation studies regrading \textbf{LoRA rank} with accuracy metric on BAPPS benchmark~\cite{bapps}. 
(b) Comparison of \textbf{different methods to distinguish multiple images}.
Metric for comparison: accuracy within Traditional / CNN categories.
Metric for reasoning: reasonable rate. 
The performance of image embedding and unique projector is obviously inferior to unique tag. 
}
\begin{minipage}[t]{0.47\textwidth}
\setlength\tabcolsep{4pt}
\scriptsize
(a) Different LoRA ranks. 
\begin{flushleft}
\begin{tabular}{c|ccccc}
\toprule
LoRA Rank  & 8    & 12   & 16            & 24   & 32 \\
\midrule
Traditional & fail & 79.7 & 80.0 & 79.6 & 80.5 \\
CNN         & fail & 83.9 & 83.8 & 84.0 & 83.8 \\
\midrule
Average     & fail & 81.8 & 81.9 & 81.8 & 82.1 \\
\bottomrule
\end{tabular}
\label{appendix:tab:ablation_R}
\end{flushleft}
\end{minipage}
\hfill
\begin{minipage}[t]{0.51\textwidth}
\setlength\tabcolsep{5pt}
\renewcommand{\arraystretch}{1.15}
\scriptsize
(b) Various methods to distinguish multiple images. 
\begin{flushright}
\begin{tabular}{c|cc}
\toprule
Method           & Comparison  & Reasoning \\
\midrule
Image Embedding  & 67.0 / 66.9 &  26.0 \\
Unique Projector & 75.1 / 77.2 &  40.0 \\
Unique Tag       & \textbf{80.0} / \textbf{83.8} &  \textbf{53.0} \\
\bottomrule
\end{tabular}
\label{appendix:tab:ablation_multi_img}
\end{flushright}
\end{minipage}
\end{table}

\textbf{Influences of LoRA rank} are illustrated in \cref{appendix:tab:ablation_R}a. 
When LoRA rank is too small (\eg, 8), the training process diverges. 
When the training process converges with larger LoRA rank (\eg, 12, 16, 24, and 32), our \model is insensitive to this hyper-parameter.
The default choice of LoRA rank is 16.

\textbf{Different methods to distinguish multiple images}. 
In addition to the unique tag method introduced in the main paper, we investigate alternative strategies for managing multiple images. 
(1) Image embedding. 
We add a trainable embedding to visual tokens for each image type. 
In \textit{comparison} and \textit{reasoning} tasks, distinct embeddings are assigned to the reference image, Image A, and Image B. 
For the \textit{description} task, the embeddings for the reference image and Image A are added to the tokens of the reference images and distorted images, respectively. 
(2) Unique image projector. 
We utilize unique image projectors for varied image categories. 
These projectors are specifically initialized for the reference image, Image A, and Image B in \textit{comparison} and \textit{reasoning} tasks. 
In the \textit{description} task, only projectors for the reference image and Image A are employed. 
The experimental results are illustrated in \cref{appendix:tab:ablation_multi_img}b. 
The performance of image embedding and unique projector is obviously inferior to unique tag. 
We suspect that the reason could be that changing the visual tokens differently for various images may confuse the model.

\begin{figure}[t]
    \centering
    \includegraphics[width=0.8\linewidth]{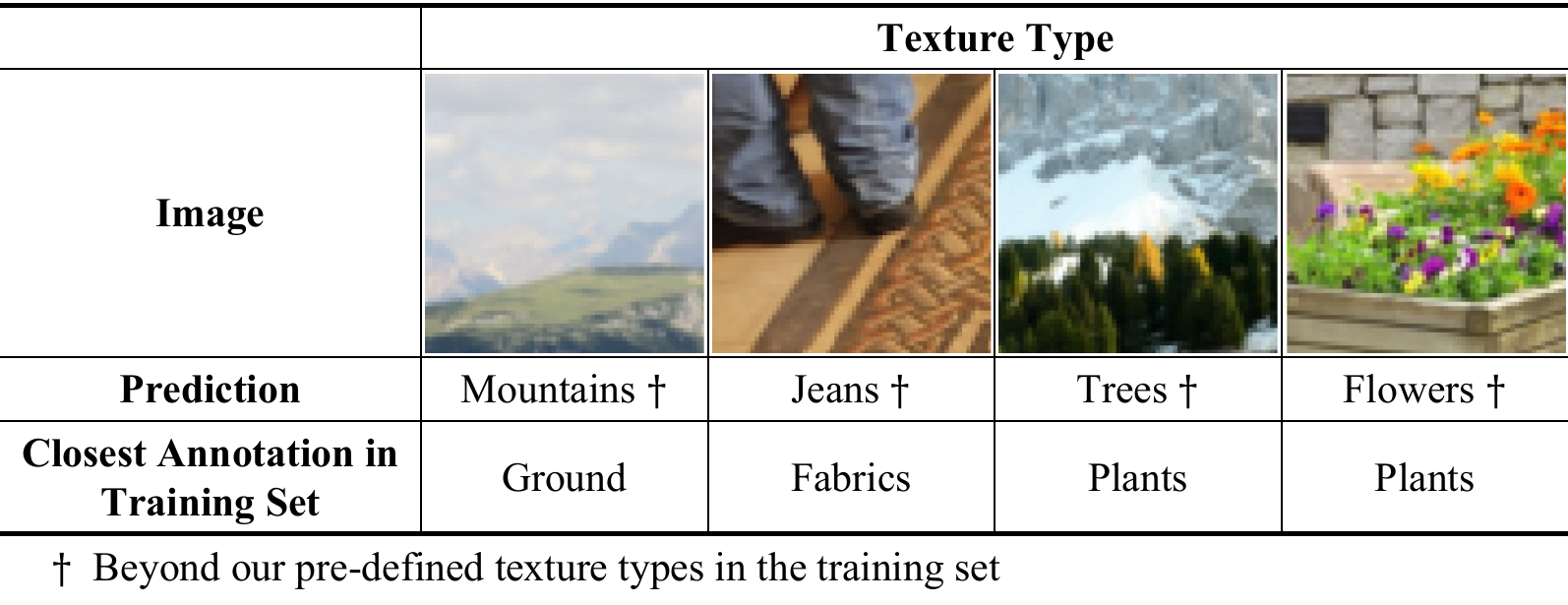}
    \caption{
        \textbf{Training with the content description data improves the diversity of texture types in responses}, which even beyond our pre-defined texture types in training set.
    }
    \label{appendix:fig:coco}
\end{figure}

\textbf{Impacts of content description data on texture diversity}. 
Incorporating external content description data, specifically the Detailed Description Dataset in~\cite{lamm}, stably enhances the performance, as discussed in the main paper. 
Training with content description data also enriches the diversity of texture types recognized, even beyond the pre-defined categories in our training set, as depicted in \cref{appendix:fig:coco}. 
For instance, in the last two examples in \cref{appendix:fig:coco}, though the texture types of ``trees'' and ``flowers'' are usually annotated as ``plants'' in the training set, we still correctly predict these fine-grained texture types.

\section{More Results}\label{appendix:sec:results}


\subsection{Results on General Multi-modal LLMs}

In the main paper, we show that general MLLMs are not capable of tackling IQA tasks. 
One possible reason for general MLLMs' inadequacy is that they are usually trained on single-image input, while our tasks involve multi-image input. 
Therefore, we also test LLaVA-1.5~\cite{llava1.5} on \textit{description} task without the reference image (\ie, single-image input). 
The qualitative results in \cref{appendix:fig:res_llava} confirm the unsatisfying results of general MLLMs even with single-image input.

\begin{figure*}[t]
    \centering
    \includegraphics[width=1.0\linewidth]{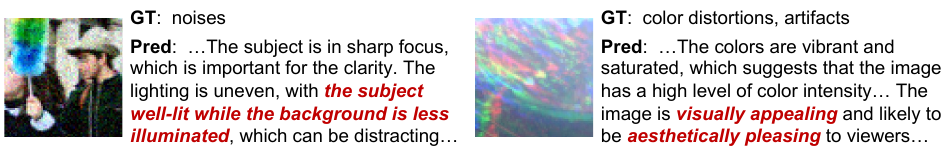}
    \caption{
        \textbf{Qualitative results of LLaVA-1.5~\cite{llava1.5} on non-reference \textit{description} task}. 
        Even with single-image input, the results is still unsatisfying. 
    }
    \label{appendix:fig:res_llava}
\end{figure*}

\subsection{More Results of Quality Description}

\textbf{Qualitative results of quality description} are illustrated in \cref{appendix:fig:res_description_0} and \ref{appendix:fig:res_description_1}. 
We sample these results from: 
\begin{itemize}
    \item Different distortions. \eg, \textit{blurriness} in \ref{appendix:fig:res_description_0}a \& \ref{appendix:fig:res_description_1}c, \textit{noise} in \ref{appendix:fig:res_description_0}c \& \ref{appendix:fig:res_description_1}e, \textit{color distortion} in \ref{appendix:fig:res_description_0}d \& \ref{appendix:fig:res_description_1}d. 
    \item Various contents. \eg, \textit{plants} in \ref{appendix:fig:res_description_0}c \& \ref{appendix:fig:res_description_0}e, \textit{architectures} in \ref{appendix:fig:res_description_0}d \& \ref{appendix:fig:res_description_1}b, \textit{people} in \ref{appendix:fig:res_description_0}f.
    \item Diverse distortion degrees. \eg, \textit{severe} in \ref{appendix:fig:res_description_0}a \& \ref{appendix:fig:res_description_1}d, \textit{mild} in \ref{appendix:fig:res_description_0}b, \ref{appendix:fig:res_description_0}d, \& \ref{appendix:fig:res_description_0}e, \textit{no distortion} in \ref{appendix:fig:res_description_0}f \& \ref{appendix:fig:res_description_1}b.
\end{itemize}

\subsection{More Results of Quality Comparison}\label{subsec:comparison}

\begin{figure*}[ht]
    \centering
    \includegraphics[width=0.95\linewidth]{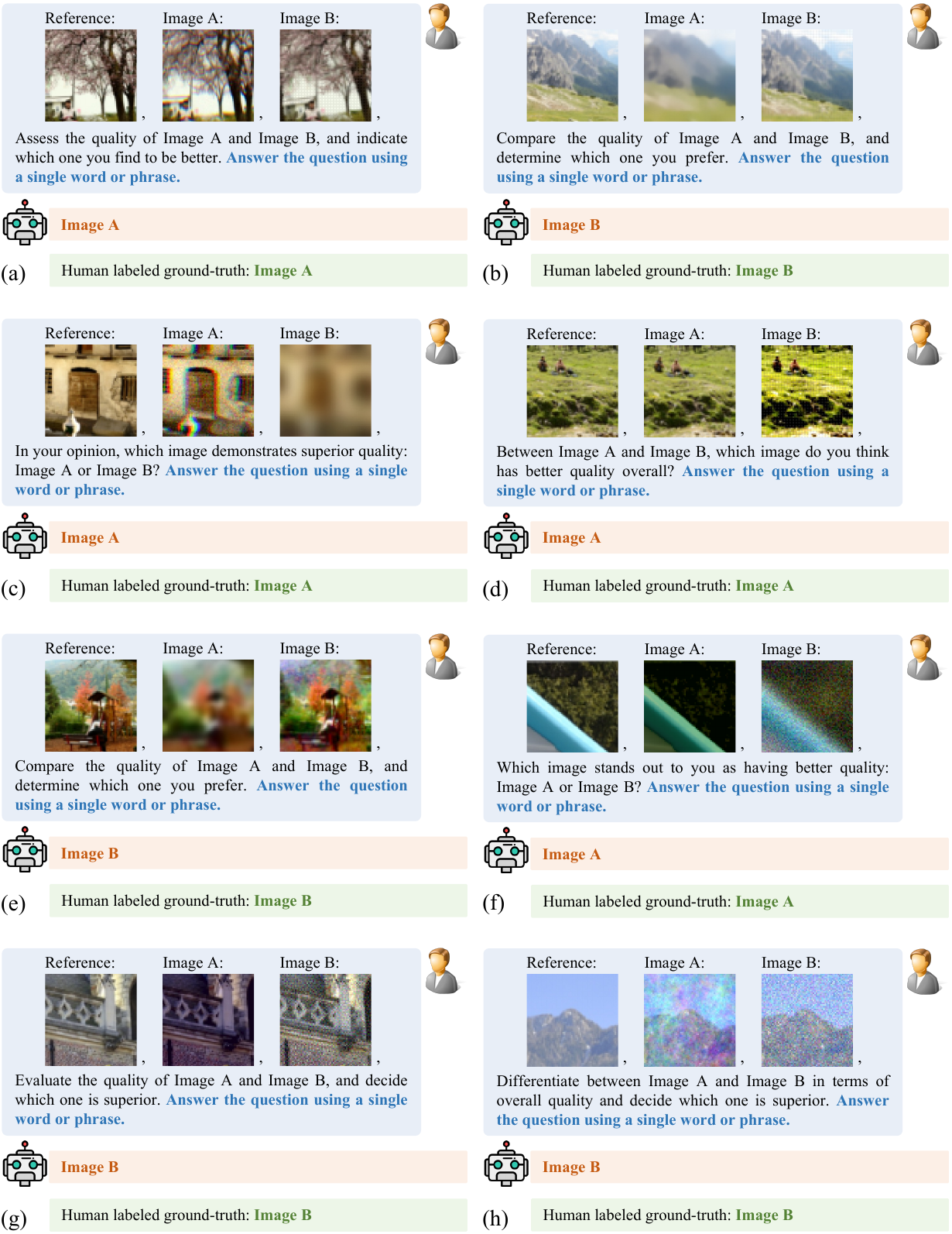}
    \caption{
        \textbf{Qualitative results of \model on quality comparison task} (with short answer prompt). 
        Our \model aligns quite well with human judgments. 
    }
    \label{appendix:fig:res_comparison_1}
\end{figure*}

\textbf{Qualitative results of quality comparison} are illustrated in \cref{appendix:fig:res_comparison_0} (without short answer prompt) and \ref{appendix:fig:res_comparison_1} (with short answer prompt). 
\model aligns closely with human judgments under the circumstances: 
\begin{itemize}
\item Different types of distortions. 
\eg, \textit{blurriness} in \ref{appendix:fig:res_comparison_1}b \& \ref{appendix:fig:res_comparison_1}c, 
\textit{noise} in \ref{appendix:fig:res_comparison_0}a \& \ref{appendix:fig:res_comparison_0}b, 
\textit{color distortion} in \ref{appendix:fig:res_comparison_0}g \& \ref{appendix:fig:res_comparison_1}f, 
\textit{artifacts} in \ref{appendix:fig:res_comparison_1}a \& \ref{appendix:fig:res_comparison_1}d. 
\item Different texture types. 
\eg, architectures in \ref{appendix:fig:res_comparison_0}a \& \ref{appendix:fig:res_comparison_1}c,
\textit{object edges} in \ref{appendix:fig:res_comparison_1}f,
\textit{water surface} in \ref{appendix:fig:res_comparison_0}b \& \ref{appendix:fig:res_comparison_0}f,
\textit{plants} in \ref{appendix:fig:res_comparison_0}d \& \ref{appendix:fig:res_comparison_1}d. 
\item Both images suffering from severe distortions.
\eg, \textit{color distortion} in A \& \textit{blurriness} in B in \ref{appendix:fig:res_comparison_0}d,
\textit{noise} in A \& \textit{blurriness} in B in \ref{appendix:fig:res_comparison_1}c. 
\end{itemize}

\begin{table}[t]
\centering
\scriptsize
\setlength\tabcolsep{8pt}
\caption{\textbf{Our \model behaves consistently with the tendencies of human annotators}. When excluding samples with high label uncertainty in BAPPS dataset~\cite{bapps}, the comparison accuracy improves significantly.}
\label{appendix:tab:margin}
\begin{tabular}{c|c|c|c}
\toprule
Excluded Labels & $\emptyset$ & $\{0.4, 0.6\}$ & $\{0.2, 0.4, 0.6, 0.8\}$ \\
\midrule
Accuracy within Traditional / CNN & 80.0 / 83.8 & 86.1 / 89.4 & 92.1 / 94.5 \\
\bottomrule
\end{tabular}
\end{table}

\textbf{Consistency with the tendencies of human annotators}. 
The ground-truth scores in the BAPPS dataset~\cite{bapps} represent the proportion of annotators favoring Image B.
In the validation set of BAPPS, there are 5 annotators in total, so the value of the ground-truth score varies across [0, 0.2, 0.4, 0.6, 0.8, 1.0]. 
As shown in \cref{appendix:tab:margin}, when we exclude the samples with high label uncertainty in the BAPPS dataset, the comparison accuracy of our \model improves significantly, indicating that our model behaves consistently with the tendencies of human annotators.

\subsection{More Results of Comparison Reasoning}

\textbf{Qualitative results of comparison reasoning} are illustrated in \cref{appendix:fig:res_reasoning_0}, \ref{appendix:fig:res_reasoning_1}, and \ref{appendix:fig:res_reasoning_2}. 
Similar to \cref{subsec:comparison}, these samples are sampled from: 
\begin{itemize}
\item Various categories of distortions. 
\eg, \textit{color distortion} in \ref{appendix:fig:res_reasoning_0}a \& \ref{appendix:fig:res_reasoning_0}c, 
\textit{blurriness} in \ref{appendix:fig:res_reasoning_1}c \& \ref{appendix:fig:res_reasoning_2}a, 
\textit{artifacts} in \ref{appendix:fig:res_reasoning_2}b \& \ref{appendix:fig:res_reasoning_2}c, 
\textit{noise} in \ref{appendix:fig:res_reasoning_1}a \& \ref{appendix:fig:res_reasoning_2}d. 
\item Different image contents. 
\eg, \textit{sky and clouds} in \ref{appendix:fig:res_reasoning_0}b, 
\textit{architectures} in \ref{appendix:fig:res_reasoning_1}a \& \ref{appendix:fig:res_reasoning_2}c, 
\textit{plants} in \ref{appendix:fig:res_reasoning_0}d \& \ref{appendix:fig:res_reasoning_1}c. 
\item Both images with severe distortions. 
\eg, \textit{color distortion} in A \& \textit{blurriness} in B in \ref{appendix:fig:res_reasoning_2}a, 
\textit{color distortion} in A \& \textit{noise} in B in \ref{appendix:fig:res_reasoning_0}c. 
\end{itemize}

\subsection{Failure Cases}

\textbf{Failure cases of \model} in \textit{quality description} and \textit{comparison reasoning} tasks are presented in \cref{appendix:fig:res_fail}. 
(1) \cref{appendix:fig:res_fail}a and \ref{appendix:fig:res_fail}b show \model incorrectly identifying distortion types, leading to wrong quality assessments in the quality description task. 
(2) In \cref{appendix:fig:res_fail}c, \model mistakenly attributes the noise in Image A to Image B, resulting in an incorrect comparison result. 
As reflected in Fig. \textcolor{red}{4} in the main paper, the unique tag approach does not completely resolve the confusion issue. 
(3) In \cref{appendix:fig:res_fail}d, \model fails to recognize an unusual distortion (\ie, complete semantic content washout in Image B). 
This is primarily due to the absence of such distortions in the training set.

\subsection{More Results of GPT-4V}

\textbf{Qualitative results of GPT-4V}~\cite{gpt4v} in \textit{comparison reasoning} task are shown in \cref{appendix:fig:res_gpt4v}. 
GPT-4V gives quite detailed comparison reasoning results, which is the main reason for its excellent GPT4-score (see Tab. \textcolor{red}{4} in the main paper).

\textbf{Failure cases of GPT-4V} in \textit{comparison reasoning} task are given in \cref{appendix:fig:fail_gpt4v}. 
Despite GPT-4V providing detailed reasoning texts, it mistakenly describes distortions, texture damages, and comparison results.

\subsection{Extensions}

\begin{figure}[t]
\centering
\includegraphics[width=0.85\linewidth]{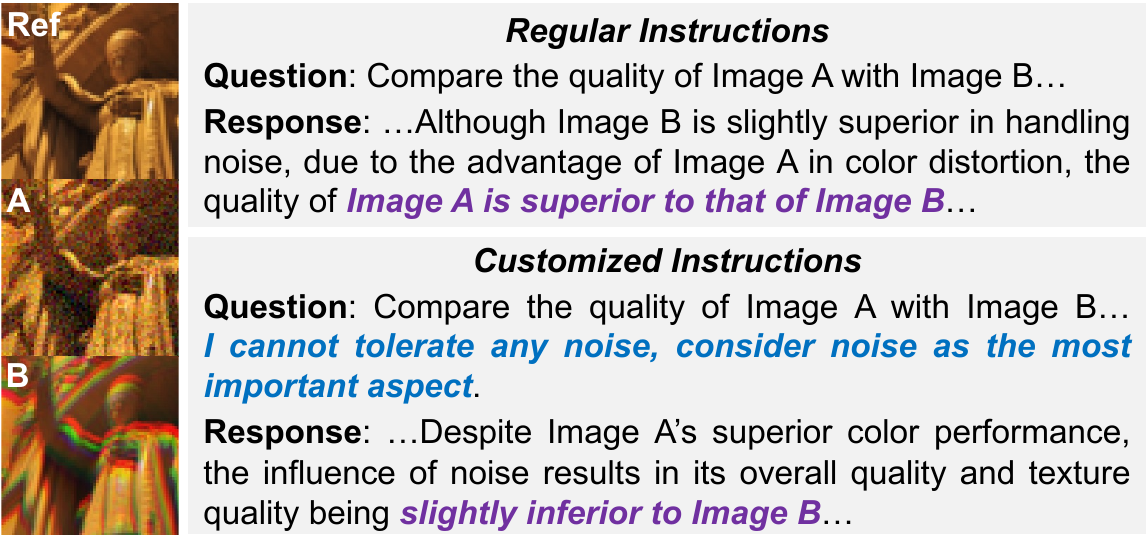}
\captionof{figure}{
    \textbf{\model has the potential to be a customized IQA tool} by incorporating human preferences into instructions. 
}
\label{appendix:fig:prompts}
\end{figure}

\definecolor{highlight_blue}{rgb}{0.0, 0.44, 0.75}
\begin{figure}[t]
    \centering
    \includegraphics[width=0.85\linewidth]{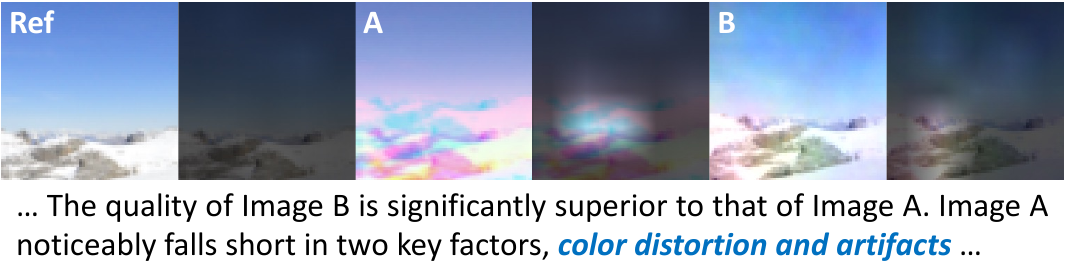}
    \caption{
        \textbf{Visualization of the attention map}. 
        We visualize decoder attention weights between the \textcolor{highlight_blue}{\textbf{highlighted words}} and the visual tokens. 
        Our model focuses on relevant regions containing text-described distortions. 
    }
    \label{appendix:fig:attention}
\end{figure}

\textbf{Potential to be customized}. 
Given various instructions, \model has the potential to be a customized IQA tool. 
To demonstrate this potential, as shown in \cref{appendix:fig:prompts}, we append the specific human preference (\eg, cannot tolerate any noise) to the instructions, \model adapts its comparison results accordingly. 
How to better induce this ability is our future work.

\textbf{Potential for distortion localization}. 
In \cref{appendix:fig:attention}, we visualize decoder attention weights between the texts describing distortions and the visual tokens. 
Our model focuses on regions containing text-described distortions. 
This invites our future research into precise distortion localization based on textual outputs.

\definecolor{color_distortion}{rgb}{0.02, 0.20, 1.00}
\definecolor{color_textue}{rgb}{0.58, 0.13, 0.57}
\definecolor{color_conclusion}{rgb}{0.0, 0.56, 0.32}

\begin{figure*}[b]
    \centering
    \includegraphics[width=0.95\linewidth]{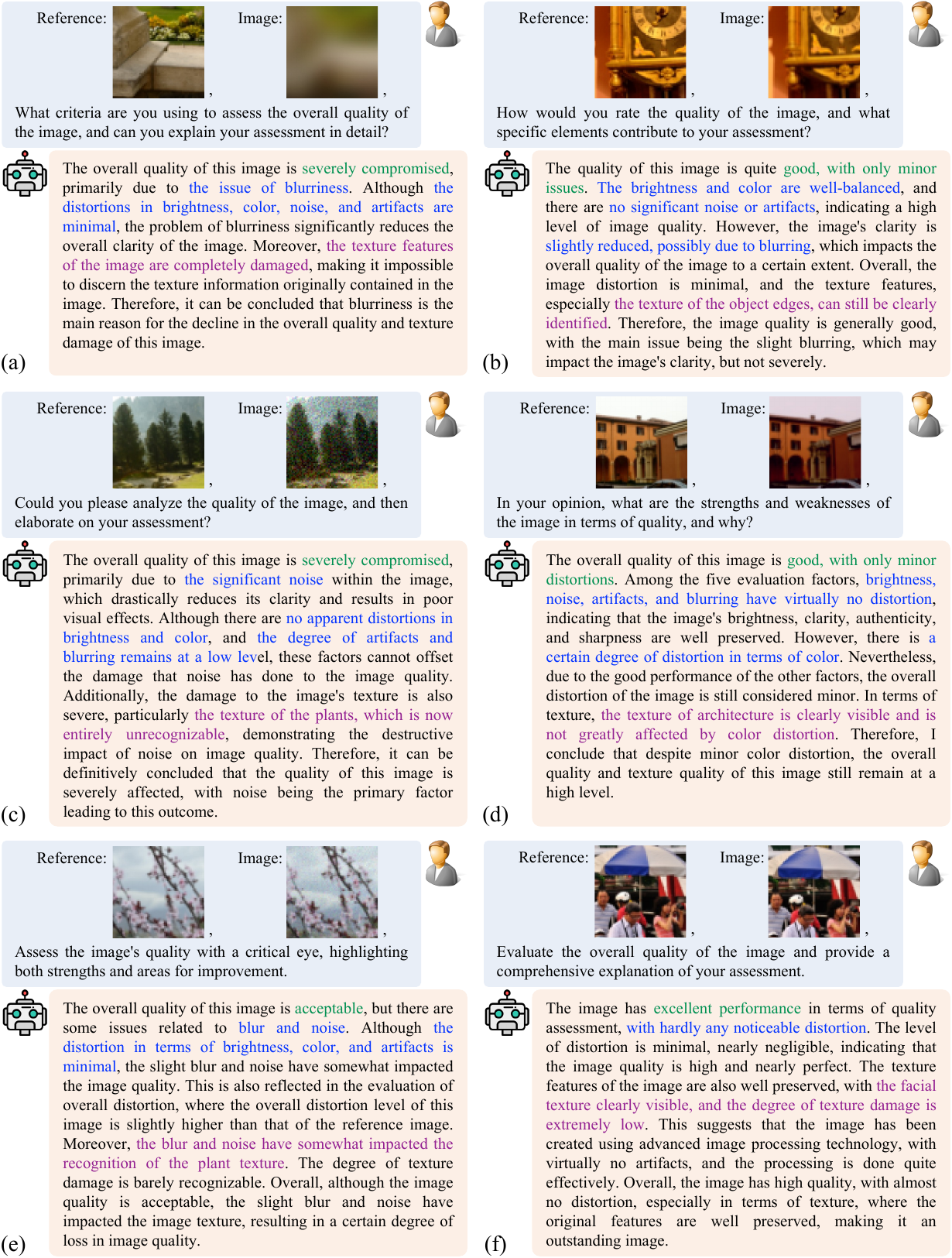}
    \caption{
        \textbf{Qualitative results of \model on quality description task}. 
        Our \model could clearly identify \textcolor{color_distortion}{distortions} and \textcolor{color_textue}{texture damages}, then weigh the impacts of various aspects to draw the final \textcolor{color_conclusion}{conclusion}. 
    }
    \label{appendix:fig:res_description_0}
\end{figure*}

\begin{figure*}[ht]
    \centering
    \includegraphics[width=0.95\linewidth]{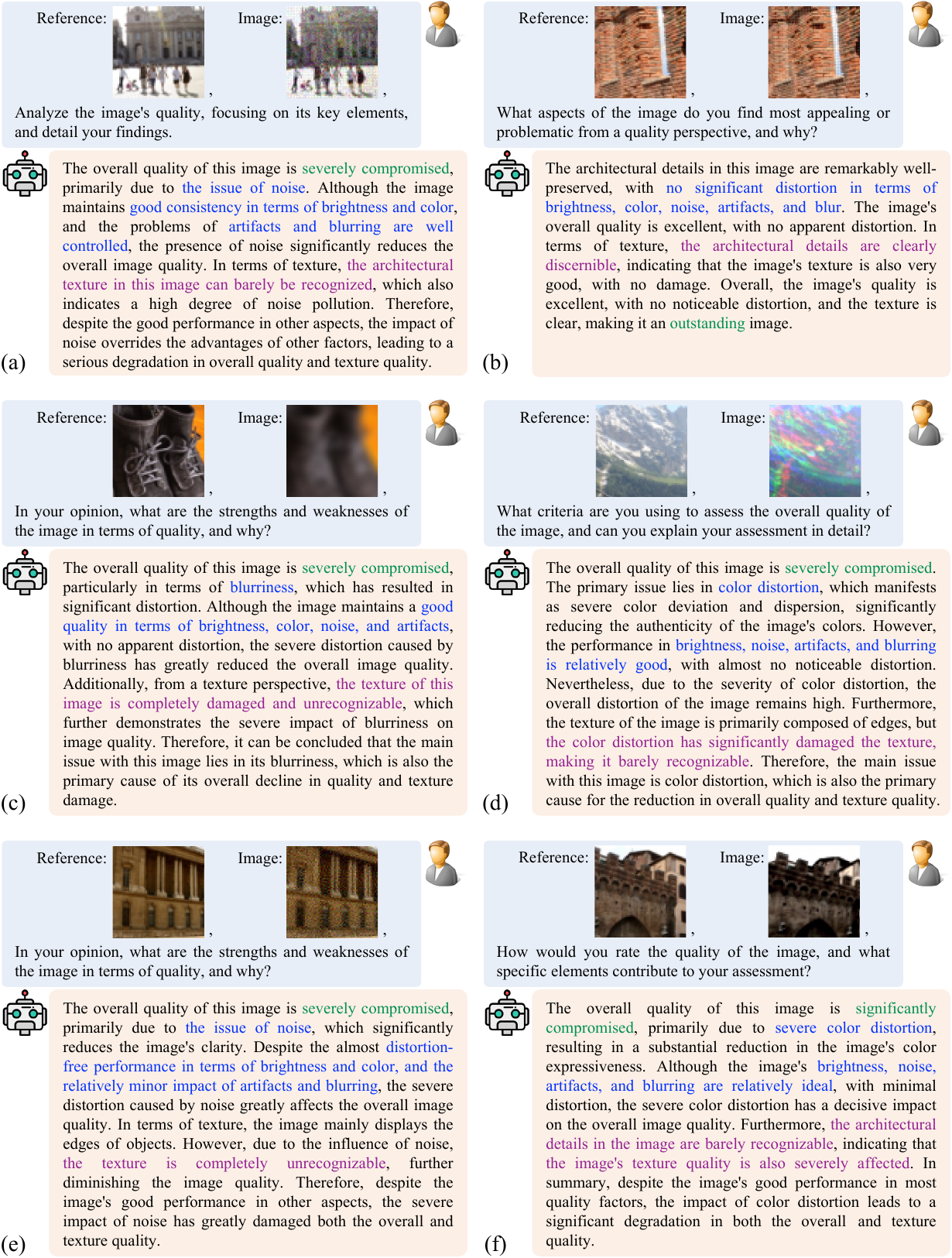}
    \caption{
        \textbf{Qualitative results of \model on quality description task}. 
        Our \model could clearly identify \textcolor{color_distortion}{distortions} and \textcolor{color_textue}{texture damages}, then weigh the impacts of various aspects to draw the final \textcolor{color_conclusion}{conclusion}. 
    }
    \label{appendix:fig:res_description_1}
\end{figure*}

\begin{figure*}[ht]
    \centering
    \includegraphics[width=0.95\linewidth]{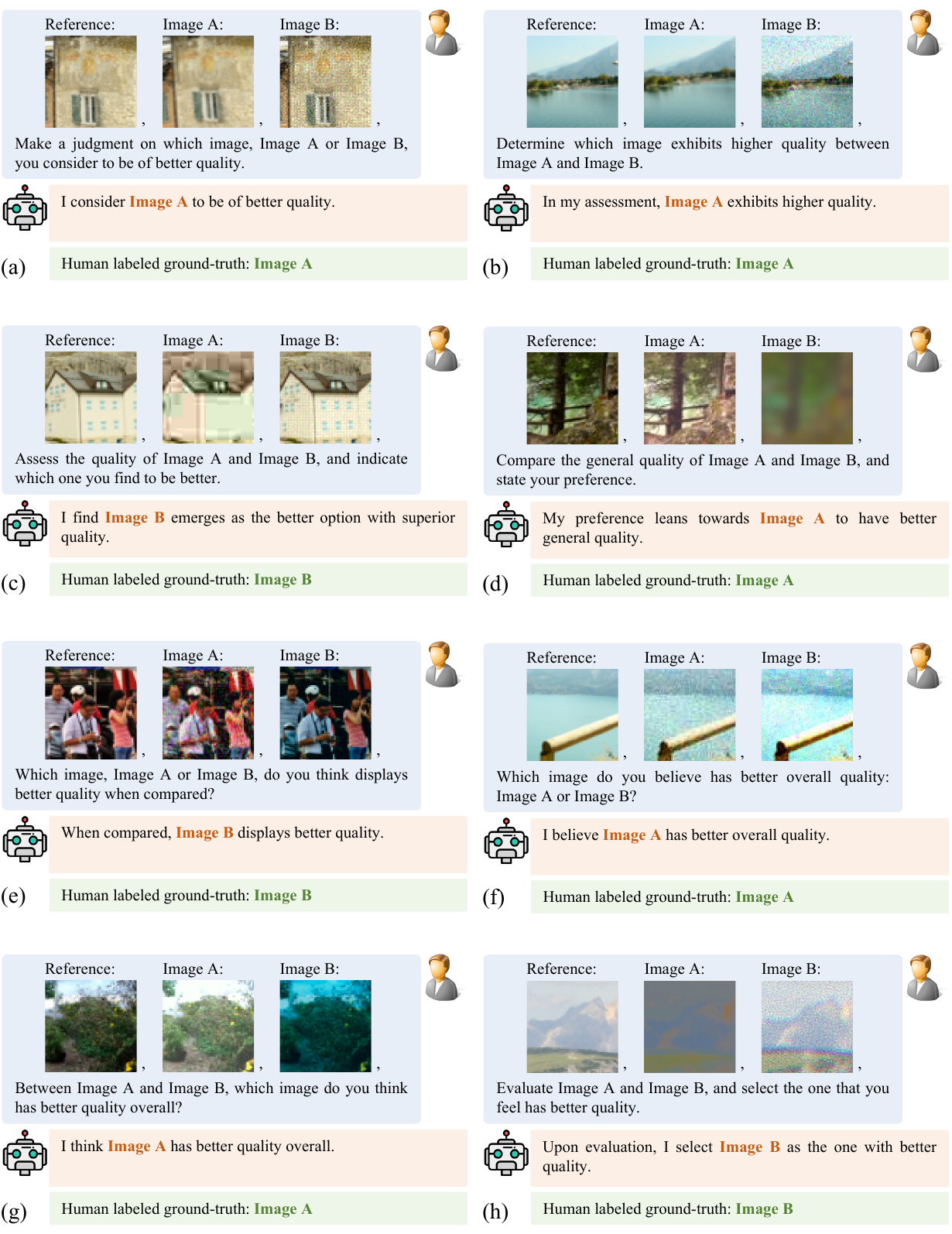}
    \caption{
        \textbf{Qualitative results of \model on quality comparison task} (without short answer prompt). 
        Our \model aligns quite well with human judgments. 
    }
    \label{appendix:fig:res_comparison_0}
\end{figure*}

\begin{figure*}[ht]
    \centering
    \includegraphics[width=0.95\linewidth]{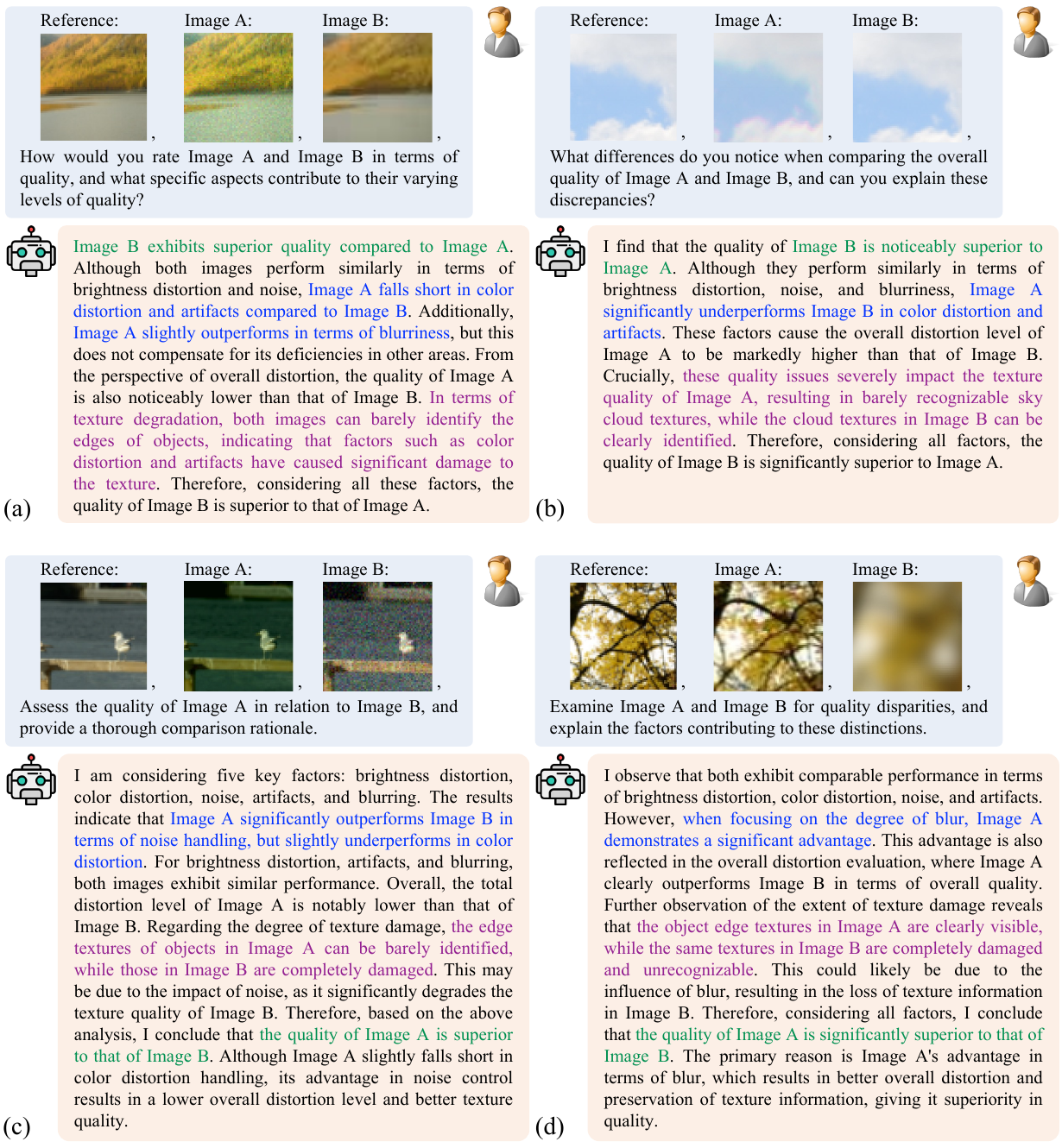}
    \caption{
        \textbf{Qualitative results of \model on comparison reasoning task}. 
        Our \model could comparatively identify \textcolor{color_distortion}{distortions}, then weigh the impacts of these distortions to the \textcolor{color_textue}{texture damages}, and finally draw the final \textcolor{color_conclusion}{comparison conclusion}. 
    }
    \label{appendix:fig:res_reasoning_0}
\end{figure*}

\begin{figure*}[ht]
    \centering
    \includegraphics[width=0.95\linewidth]{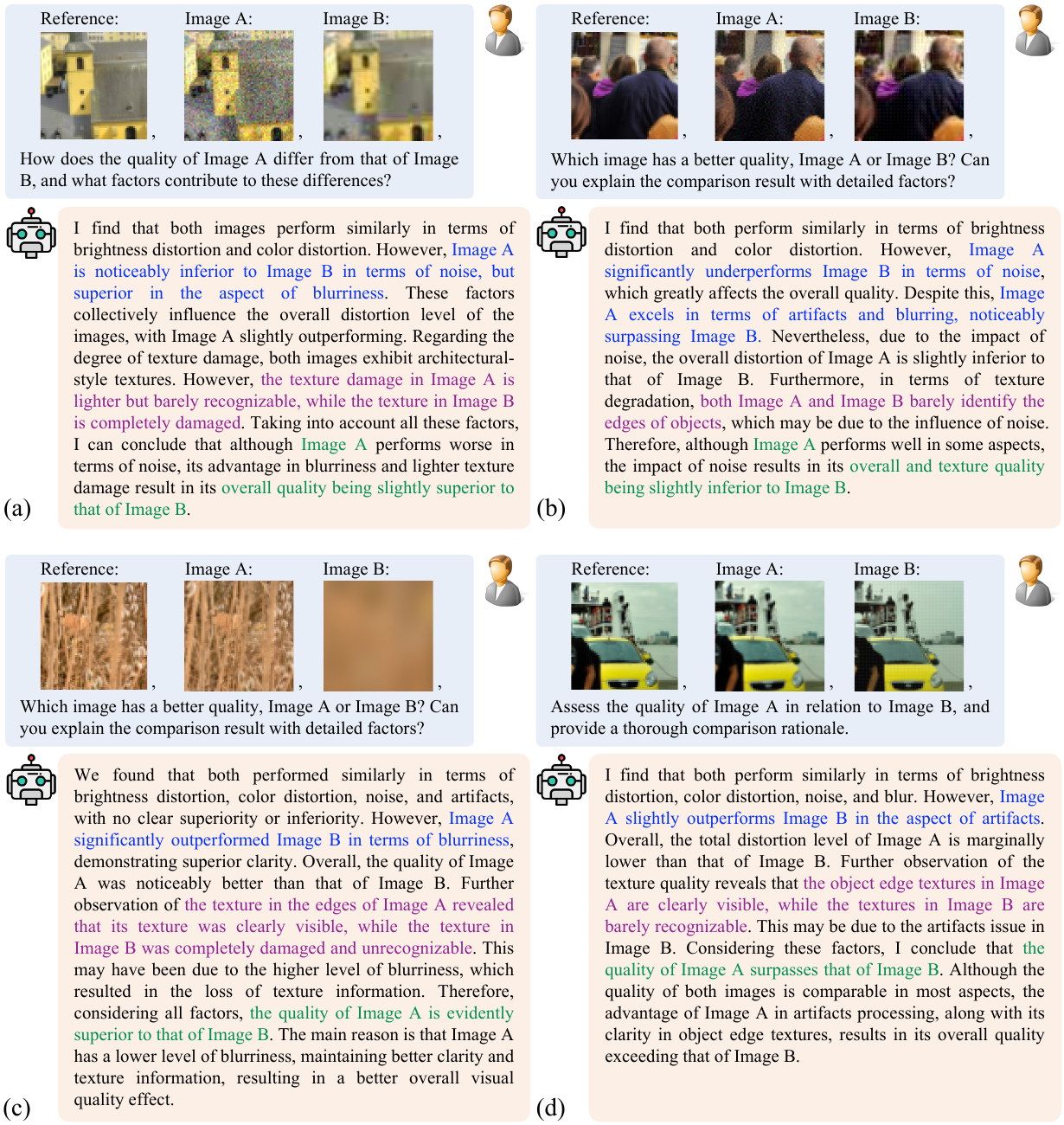}
    \caption{
        \textbf{Qualitative results of \model on comparison reasoning task}. 
        Our \model could comparatively identify \textcolor{color_distortion}{distortions}, then weigh the impacts of these distortions to the \textcolor{color_textue}{texture damages}, and finally draw the final \textcolor{color_conclusion}{comparison conclusion}. 
    }
    \label{appendix:fig:res_reasoning_1}
\end{figure*}

\begin{figure*}[ht]
    \centering
    \includegraphics[width=0.95\linewidth]{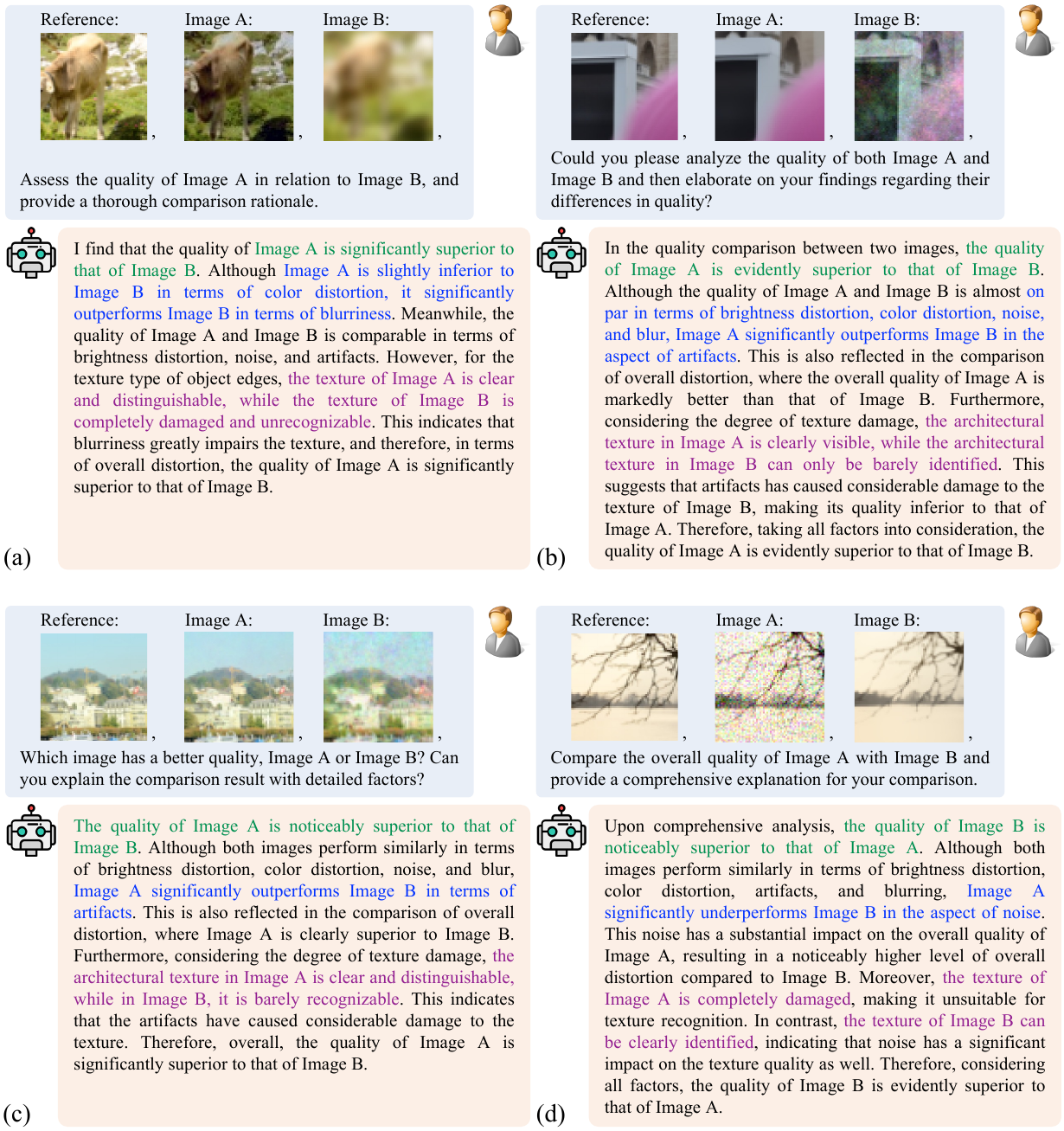}
    \caption{
        \textbf{Qualitative results of \model on comparison reasoning task}. 
        Our \model could comparatively identify \textcolor{color_distortion}{distortions}, then weigh the impacts of these distortions to the \textcolor{color_textue}{texture damages}, and finally draw the final \textcolor{color_conclusion}{comparison conclusion}. 
    }
    \label{appendix:fig:res_reasoning_2}
\end{figure*}

\begin{figure*}[ht]
    \centering
    \includegraphics[width=0.95\linewidth]{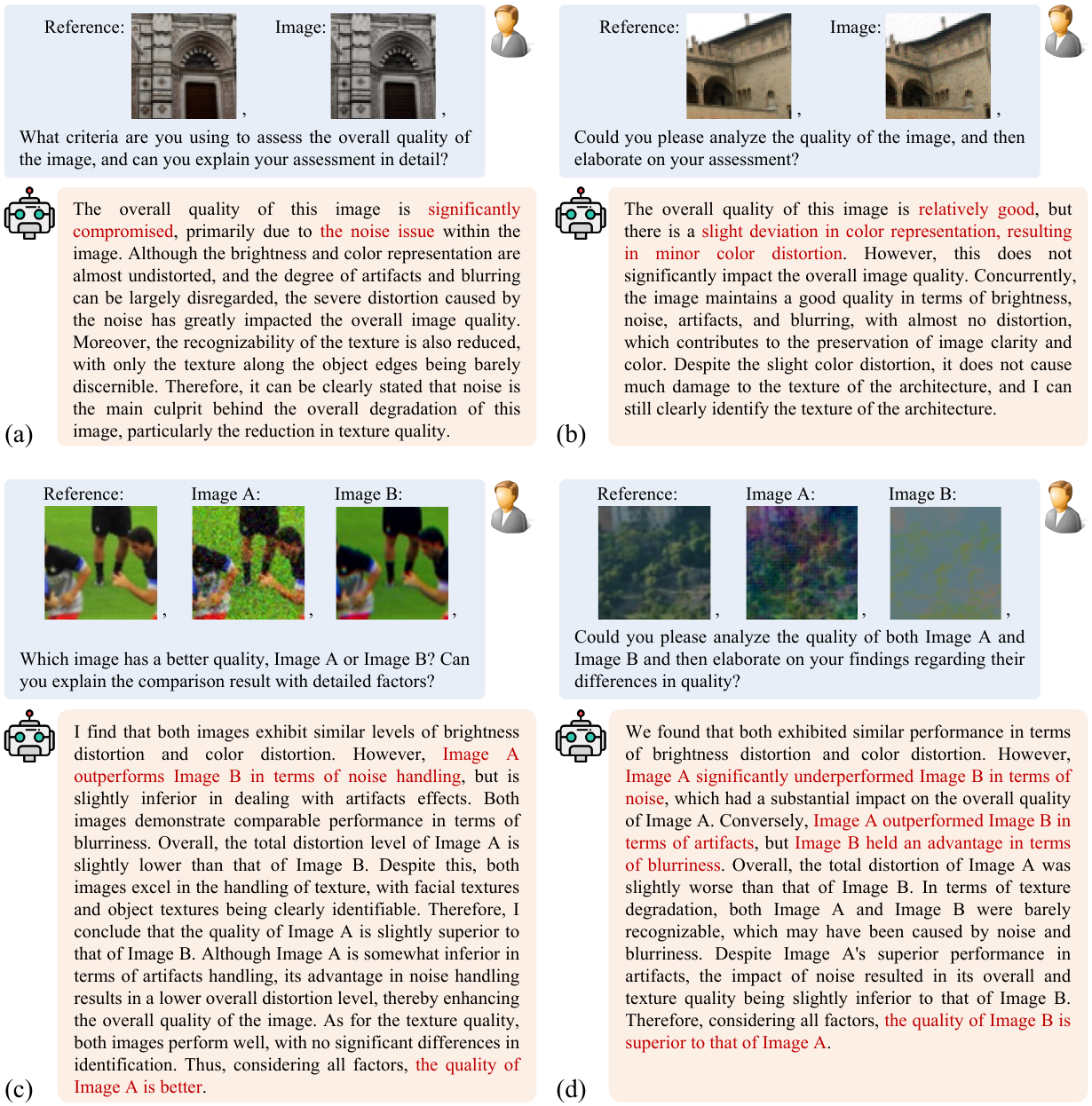}
    \caption{
        \textbf{Failure cases of \model} in \textbf{(a)(b)} quality description task and \textbf{(c)(d)} comparison reasoning task. 
    }
    \label{appendix:fig:res_fail}
\end{figure*}

\begin{figure*}[ht]
    \centering
    \includegraphics[width=0.95\linewidth]{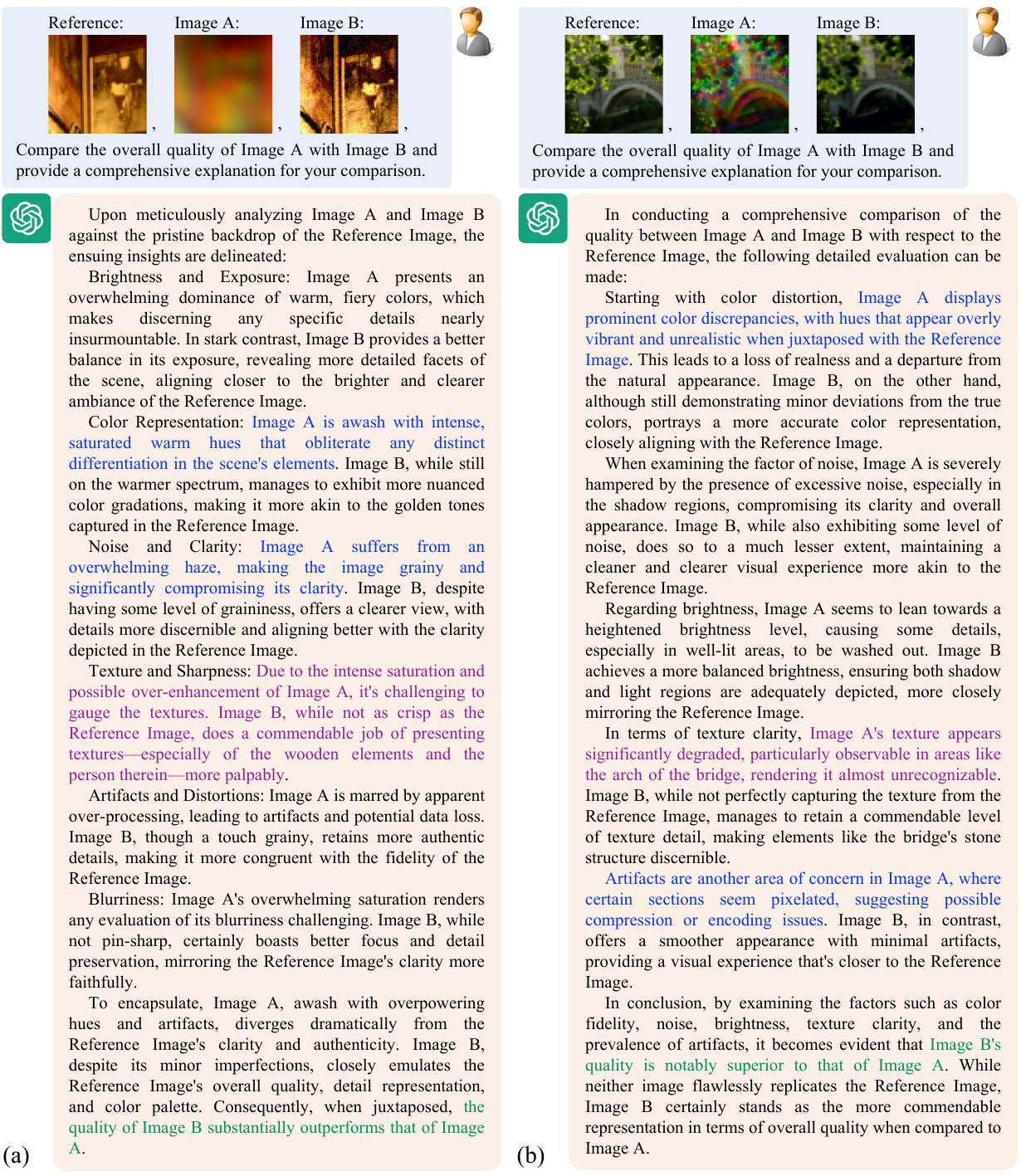}
    \caption{
        \textbf{Qualitative results of GPT-4V} in comparison reasoning task. 
    }
    \label{appendix:fig:res_gpt4v}
\end{figure*}

\begin{figure*}[ht]
    \centering
    \includegraphics[width=0.95\linewidth]{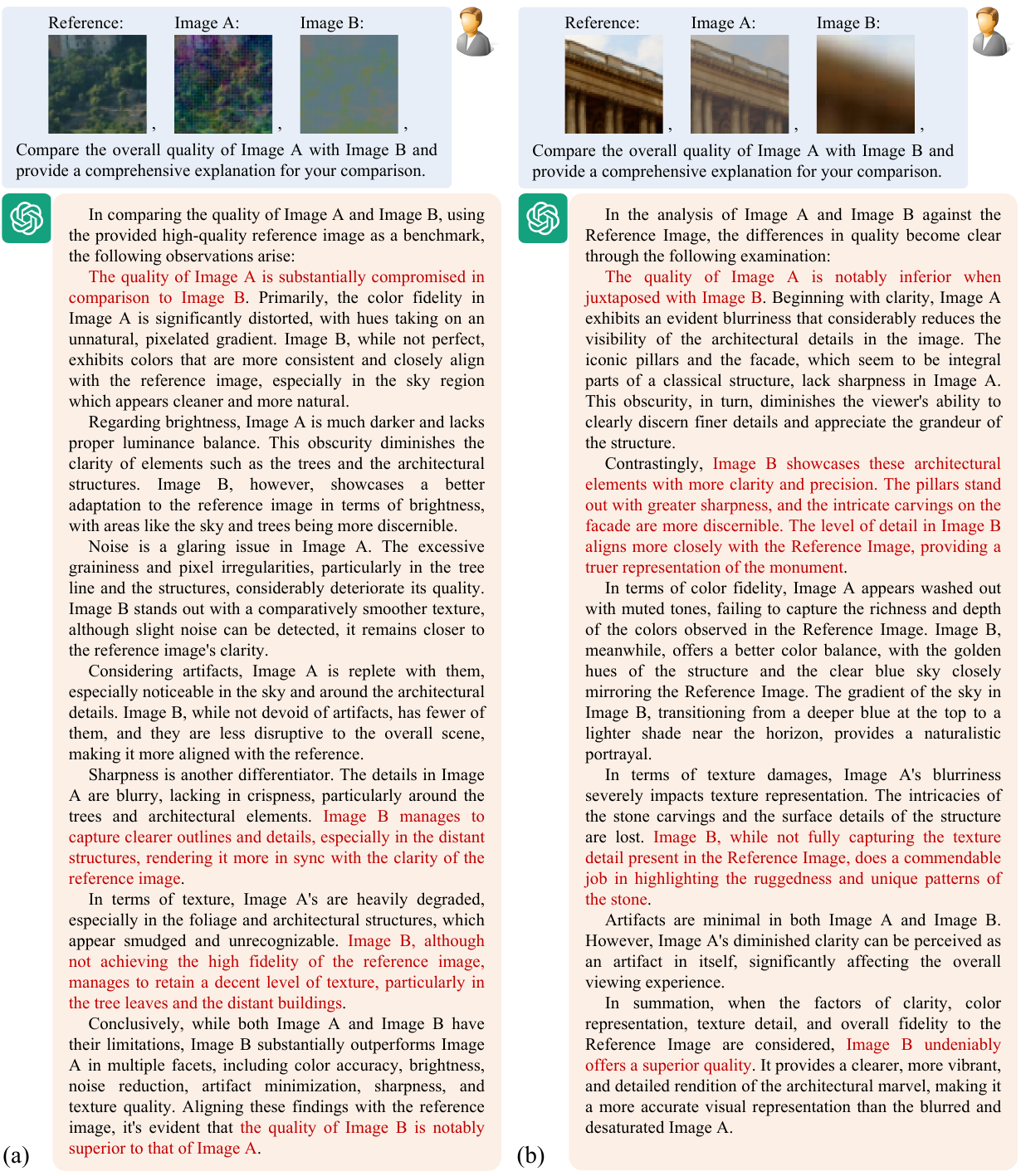}
    \caption{
        \textbf{Failure cases of GPT-4V} in comparison reasoning task. 
    }
    \label{appendix:fig:fail_gpt4v}
\end{figure*}

\end{document}